\pdfoutput=1
\documentclass[11pt]{article}

\usepackage[final]{acl}

\usepackage{times}
\usepackage{latexsym}
\usepackage[T1]{fontenc}
\usepackage[utf8]{inputenc}
\usepackage{microtype}
\usepackage{inconsolata}
\usepackage{graphicx}

\usepackage{hyperref}       % hyperlinks
\usepackage{url}            % simple URL typesetting
\usepackage{booktabs}       % professional-quality tables
\usepackage{amsfonts}       % blackboard math symbols
\usepackage{nicefrac}       % compact symbols for 1/2, etc.
\usepackage{xcolor}         % colors
\usepackage{algorithm}
\usepackage{algorithmic}
\usepackage{wrapfig}
\usepackage{caption}
\usepackage{subcaption}
\usepackage{tablefootnote}
\usepackage{listings}

\usepackage{amsmath}
\usepackage{amssymb}
\usepackage{mathtools}
\usepackage{amsthm}
\usepackage[mathscr]{euscript}

\usepackage[capitalize,noabbrev]{cleveref}

\theoremstyle{plain}
\newtheorem{theorem}{Theorem}[section]
\newtheorem{proposition}[theorem]{Proposition}

\theoremstyle{definition}

\theoremstyle{remark}

\usepackage[textsize=tiny]{todonotes}

\title{LiPO: Listwise Preference Optimization through Learning-to-Rank}

\author{
  Tianqi Liu$^\dagger$$^*$, Zhen Qin$^\dagger$$^*$, Junru Wu$^\dagger$, Jiaming Shen$^\dagger$,
  Misha Khalman$^\dagger$,
  Rishabh Joshi$^\dagger$,
  Yao Zhao$^\dagger$, \\
  \bf{Mohammad Saleh$^\dagger$,
  Simon Baumgartner$^\dagger$,
  Jialu Liu$^\ddagger$,
  Peter J. Liu$^\dagger$,
  Xuanhui Wang$^\dagger$ }\\
  \normalfont{Google DeepMind$^\dagger$, Google$^\ddagger$} \\
  \texttt{\{tianqiliu,zhenqin\}@google.com}
}

\begin{document}
\maketitle
\def\thefootnote{*}\footnotetext{Equal Contribution}\def\thefootnote{\arabic{footnote}}

\begin{abstract}
Aligning language models (LMs) with curated human feedback is critical to control their behaviors in real-world applications. Several recent policy optimization methods, such as DPO and SLiC, serve as promising alternatives to the traditional Reinforcement Learning from Human Feedback (RLHF) approach.
In practice, human feedback often comes in a format of a ranked list over multiple responses to amortize the cost of reading prompt. Multiple responses can also be ranked by reward models or AI feedback. There lacks such a thorough study on directly fitting upon a list of responses. In this work, we formulate the LM alignment as a \textit{listwise} ranking problem and describe the LiPO framework, where the policy can potentially learn more effectively from a ranked list of plausible responses given the prompt. This view draws an explicit connection to Learning-to-Rank (LTR), where most existing preference optimization work can be mapped to existing ranking objectives. Following this connection, we provide an examination of ranking objectives that are not well studied for LM alignment, with DPO and SLiC as special cases when list size is two. In particular, we highlight a specific method, LiPO-$\lambda$, which leverages a state-of-the-art \textit{listwise} ranking objective and weights each preference pair in a more advanced manner. We show that LiPO-$\lambda$ can outperform DPO variants and SLiC by a clear margin on several preference alignment tasks with both curated and real rankwise preference data. 
\end{abstract}

\section{Introduction}
\label{sec:intro}

Recent Large Language Models, such as GPT-4~\citep{openai2023gpt4} and Gemini~\citep{team2023gemini}, have unlocked unprecedented capabilities, witnessed by impressive performance on diverse tasks from conversational chatbot to programming. A key step to control the behavior of such Language Models (LMs) is to align them with curated human feedback. Reinforcement Learning with Human Feedback (RLHF)~\citep{rlhf} was first introduced to improve the alignment of LMs with human preferences~\citep{ouyang2022training}. However, RLHF is a complex process, requiring substantial memory and hyperparamter tuning.

\begin{figure}
% https://colab.corp.google.com/drive/1yyDOgLn6goobILuukMb9c2tOoLywckDf?usp=sharing
% \begin{figure}
  \centering
  \includegraphics[width=0.4\textwidth]{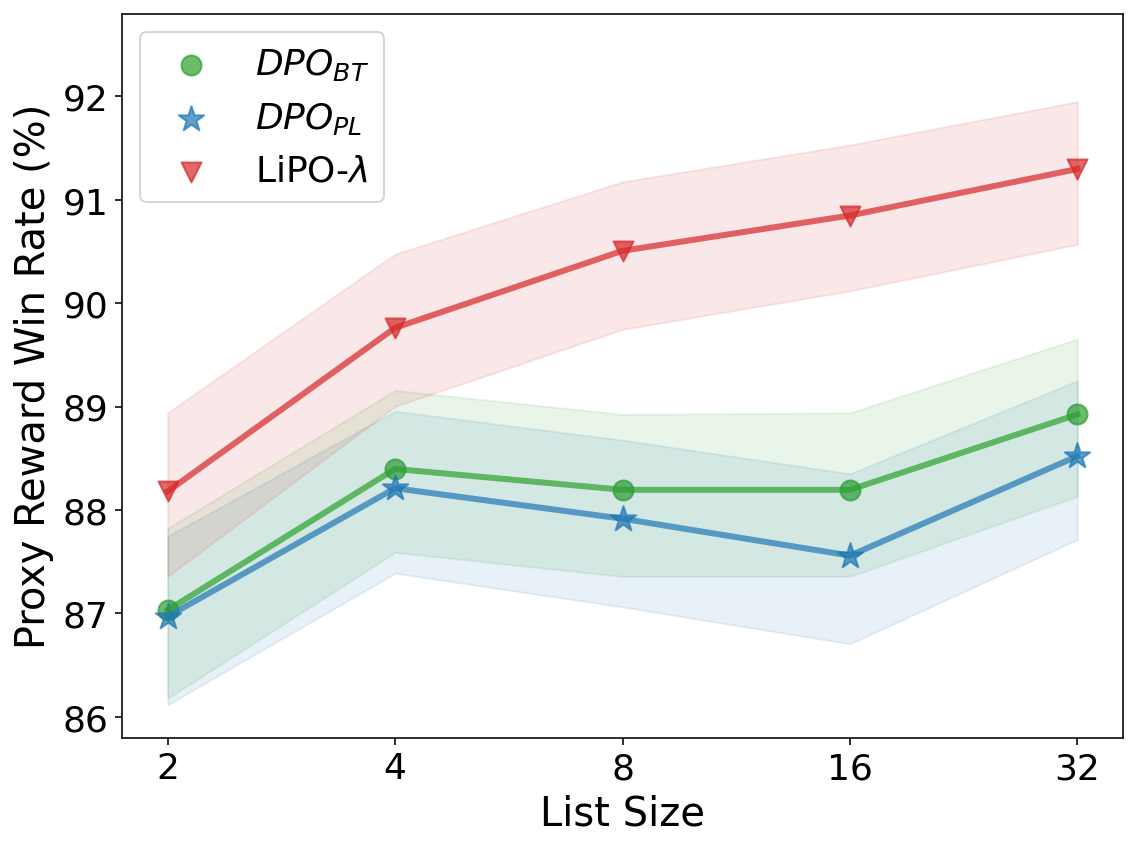}
  \caption{Performance of the pairwise DPO (DPO$_{\textrm{BT}}$), listwise DPO (DPO$_{\textrm{PL}}$), and our new listwise approach LiPO-$\lambda$. All benefit from training data beyond pairwise data (List Size = 2), while LiPO-$\lambda$ can benefit more and monotonically as list size increases. 95\% bootstrapped confidence intervals are shown as shaded areas.}  
  \label{figure:teaser}
% \end{figure}
\end{figure}

Several recent works resort to alternatives of RLHF, and noticeably converge to a pairwise ranking optimization paradigm. For example, DPO~\citep{dpo} optimizes a pairwise logistic loss directly from pairwise human preference data, while avoiding an explicit reward model and RL-based optimization. Similarly, SLiC~\citep{slic} optimizes a pairwise hinge loss objective on pairwise preference data directly from human or reward model ranked samples from the supervised fine-tuned (SFT) policy. RRHF~\citep{rrhf} starts from listwise preference data labeled by a reward model, and optimizes a pairwise contrastive objective by comparing all pairs in the list, which is analogous to the SLiC objective. 
%More recently, RSO~\citep{rso} unifies DPO and SLiC losses and proposes a better way of sourcing preference data via statistical rejection sampling. 

Virtually all popular preference optimization frameworks do not go beyond pairwise preferences. However, in practice, human preference data can come as a ranked list to amortize the cost of reading prompt~\citep{kopf2024openassistant,ouyang2022training}. It is also highly efficient for sampling a list of responses from a shared prompt~\citep{pope2023efficiently}. These motivate us to study preference optimization on listwise data. Some recent work did study preference optimization using listwise ranking objectives. For example, the DPO paper~\citep{dpo} briefly touches the listwise Plackett-Luce model~\citep{luce2005individual} for preference modeling without any experimental results. Meanwhile, PRO~\citep{song2023preference}, studied the Plackett-Luce model more formally. However, existing work are usually ad-hoc, and there lacks a thorough study of the listwise ranking perspective for the LM preference optimization problem, which will bring in new insights and solutions as shown in this paper. 

In this work, we formally formulate LM alignment as a listwise ranking problem, where the LM can potentially learn alignment more effectively from listwise preferences. This for the first time draws an explicit connection to the rich Learning-to-Rank (LTR) literature~\citep{8186875}. Noticeably, the LTR literature has shown that a direct listwise optimization can be more effective than pairwise alternatives for the listwise ranking problem, with a rich set of methods of various properties and performances. In particular, popular existing methods can be mapped to existing ranking objectives. As we will show, \textit{not every listwise objective is effective} and thus it is desired to conduct an close examination to identify the most effective ones for LM alignment. In Figure~\ref{figure:teaser}, we can see that listwise preference \textit{data} can benefit existing pairwise methods like DPO even though they treat all pairs from a list equally, a problem not well studied in the literature, while our new listwise method, inspired by our general formulation, called LiPO-$\lambda$, can further benefit from listwise preferences.

% By examining them from the lens of LTR, we notice that their approach maps to a specific listwise ranking objective~\citep{xia2008listwise}, which falls behind the state of the art. 

To this end, we provide the first comprehensive study of ranking objectives under the listwise preference optimization (LiPO) framework, which allows us to compare popular and state-of-the-art ranking objectives for the LM preference optimization problem. In particular, existing popular methods, such as DPO, RRHF, and SLiC, can be mapped to existing pairwise ranking optimization objectives from listwise data (if only pairwise data is used, it is equivalent to list size being 2). On the other hands, existing listwise methods, such as PRO, usually map to \textit{ineffective} listwise objectives and LiPO allows us to analyze their pitfalls. Furthermore, inspired by the LiPO framework, we show that a new method, LiPO-$\lambda$, which leverages a state-of-the-art ranking objective~\citep{NIPS2006_af44c4c5, wang2018lambdaloss, jagerman2022optimizing}, can achieve very competitive performance. Noticeably, LiPO-$\lambda$ allows an intuitive interpretation: it leverages a sophisticated weighting paradigms that assigns listwise-aware weighting to sampled pairs, to optimize a well-founded ranking metric~\citep{wang2013theoretical}, in contrast to existing methods that assign uniform weights or use weighting schema that fails to consider various factors. 
%In terms of implementation, a listwise reward model is ideal but impractical - we show how we derive a robust ranking list for sampled listwise data, using a pairwise reward model, which is connected to existing effective ranking paradigms~\citep{qin2023large}. 

By a comprehensive study of various ranking objectives on Reddit TL;DR, AnthropicHH, and OpenAssistant tasks, we show that LiPO-$\lambda$ is better than existing methods, including (pairwise and listwise) DPO~\citep{dpo}, SLiC~\citep{slic}, PRO~\citep{song2023preference}, as well as other baselines motivated by the LTR literature.%, over a range of language model sizes with up to XB parameters.

Our contributions are summarized as follows:
\begin{itemize}
    \item We describe the novel Listwise Preference Optimization (LiPO) framework, which generalizes many recent preference optimization methods and allows an examination of such methods through the lens of LTR for a deeper understanding.
    \item We for the first time provide a comprehensive investigation of ranking objectives for LM preference optimization, especially listwise objectives that are not well studied in the LM preference optimization literature.
    \item We highlight a new method, inspired by the LiPO framework and advanced in the LTR literature, LiPO-$\lambda$, which shows competitive performance across the evaluation tasks.
\end{itemize}

%IPO paper shows that RLHF and DPO are suspicious to overfitting especially under determinstic preference and limited data regime. The same argument applies for the LTR perspective: limited number of pairs are \textit{inefficient} to learn listiwise objective and can be prone to overfitting under BT model. Listwise objective is more efficient while all pairs 

\begin{figure*}
  \centering
  \includegraphics[width=0.8\textwidth]{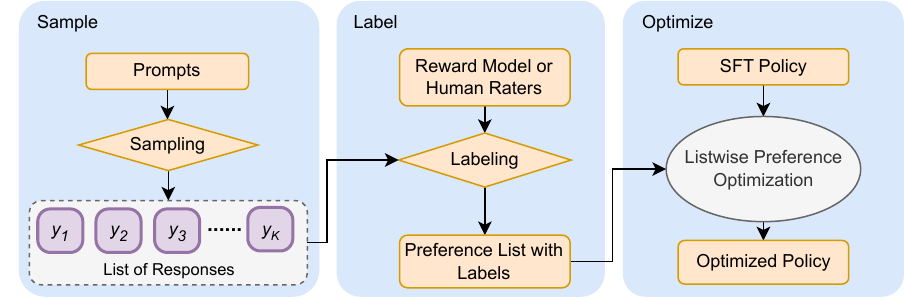}
% Feel free to modify the pipeline here: https://drive.google.com/file/d/1kKadMVnouEpUsEmc9BYb6VsCjad-9yef/view?usp=sharing
  \caption{An illustration of the Listwise Preference Optimization (LiPO) pipeline. For each prompt, LiPO samples a list of responses from certain policy or mixed policies. Then human or reward ranking model assigns each response a ranking label. After that LiPO optimizes the policy via ranking loss to align with the preferences.}  
  \label{figure:lipo_pipeline}
% \vspace{-0.8em}
\end{figure*}
\section{The LiPO Framework}
\subsection{Preliminary}
In LM generation, given a prompt, $x \in \mathcal{X}$,
there is an action space $\mathcal{Y}$, where each action is usually called a response. A policy $\pi \in \Delta^{\mathcal{X}}_{\mathcal{Y}}$ associates to each prompt $x$ a discrete probability distribution $\pi(.|x) \in \Delta_{\mathcal{Y}}$ where $\Delta_{\mathcal{Y}}$ is the set of discrete distributions over $\mathcal{Y}$. The goal is to learn a policy $\pi$ from training data, with a key consideration that the policy should align with human preference. Existing work~\citep{dpo} mainly focus on learning from \textit{pairwise} preference data.

The training dataset for DPO is $\mathcal{D} = \{x^{(i)},y_w^{(i)}, y_l^{(i)} \}_{i=1}^{N}$: given a prompt $x$, we have two responses where $y_w$ is prefered over $y_l$. The training objective of DPO (Eq.7) is 
\begin{equation}
\resizebox{1.01\hsize}{!}{$
-\mathbb{E}_{(x, y_w, y_l) \sim \mathcal{D}} \left[ \log\sigma(\beta\log{\frac{ \pi_{\theta}(y_w|x)}{ \pi_{\textrm{ref}}(y_w|x)}} - \beta\log{\frac{ \pi_{\theta}(y_l|x)}{ \pi_{\textrm{ref}}(y_l|x)}})\right],$
}
\end{equation}
%\vspace{-0.3em}
where $\pi_{\theta}$ is the policy parameterized by $\theta$ that is to be learned, $\pi_{\textrm{ref}}$ is the SFT policy that initializes $\pi_{\theta}$, and $\beta$ is a hyper-parameter to control the KL divergence between $\pi_{\theta}$ and $\pi_{\textrm{ref}}$. \citet{dpo} showed that $\beta\log{\frac{ \pi_{\theta}(y|x)}{ \pi_{\textrm{ref}}(y|x)}}$ can be treated as ``implicit reward'', the goal of DPO is to align the ``implicit reward'' towards human preference data directly, which is shown to be effective.

\subsection{The Listwise Formulation}\label{sec:formulation}

We note that preference modeling can be treated as a more general \textit{listwise} ranking problem: eventually the policy $\pi$ is expected to virtually rank a list of discrete actions from $\Delta_{\mathcal{Y}}$, and learning from listwise preference data may be more effective for LM alignment. As discussed above, human preference data can come as a ranked list to amortize the cost of reading the prompt~\citep{kopf2024openassistant,ouyang2022training}. Thus, we describe a general Listwise Preference Optimization framework. 

In the listwise formulation, the training dataset is $\mathcal{D} = \{x^{(i)}, \mathbf{y}^{(i)}, \mathbf{\psi}^{(i)}\}_{i=1}^{N}$: given a prompt $x$, we have a \textit{list} of responses $\mathbf{y} = (y_1, ..., y_K)$ of size $K$, which can be generated from SFT policy or other sources. When $K=2$, this reduces to pairwise data. We assume there are real-valued labels $\mathbf{\psi} = (\psi_1, ..., \psi_K)\in [0,1]^K$ associated with the corresponding responses, which may come from human raters or be derived from a reward model. A higher label value indicates a better response.  
\subsection{LM alignment as Learning-to-Rank}
Given the listwise preference data, LM alignment can be formulated as a Learning-to-Rank (LTR) problem. In LTR~\citep{8186875}, the goal is to learn a ranking model $\pi_{\theta}$ that can output the relevance scores $\mathbf{s}$ for all documents given a query.
In LM alignment, we treat $x$ as the query and $\mathbf{y}$ as documents in the LTR setting. Then we define the scores $\mathbf{s}=\{s_1, ..., s_K\}$, where $s_i$ is defined as the following normalized one for $(x, y_i)$ inspired by~\citet{dpo}: $\mathbf{s}(\pi_\theta, \pi_\text{ref}, \beta) = \{s_1, ..., s_K\}\triangleq$
\begin{equation}
\label{eq:scores}
     \resizebox{0.8\hsize}{!}{$  \left\{\beta\log{\frac{ \pi_{\theta}(y_1|x)}{ \pi_{\textrm{ref}}(y_1|x)}}, ..., \beta\log{\frac{ \pi_{\theta}(y_K|x)}{ \pi_{\textrm{ref}}(y_K|x)}}\right\},$}
\end{equation}
To simplify the notation, we omit the dependency of $\mathbf{s}$ upon $(\pi_\theta, \pi_\text{ref}, \beta)$ from now on.

LTR algorithms learn $\pi_{\theta}$ using loss functions as their objectives. A ranking loss function is in general defined based on labels $\mathbf{\psi}$ of responses $\mathbf{y}$ and predicted scores $\mathbf{s}$:
\begin{equation}
\label{eq:lipo}
 \mathcal{L}(\pi_{\theta};\pi_\textrm{ref},\beta) = \mathbb{E}_{(x, \mathbf{y}, \mathbf{\psi}) \sim \mathcal{D}} \left[ l(\mathbf{\psi}, \mathbf{s})\right].  
\end{equation}
$l$ is the loss function for a single prompt $x$ that takes the labels and scores as input and output a real value as the loss:
\begin{equation}
    l: (\mathbf{\psi}, \mathbf{s}) \rightarrow \mathbb{R}.
\end{equation}
A Learning-to-Rank algorithm is to find the optimal $\pi_{\theta}$ that minimizes the overall loss in the space of ranking models. We call Eq~\ref{eq:lipo} the LiPO framework under the context of LM alignment.

\subsection{Ranking Losses in Existing Work}

With the definition of $\mathbf{s}$ in Eq~\ref{eq:scores}, we show that several popular LM alignment methods can be mapped into the LiPO framework using different ranking losses. Proofs of the propositions below can be found at Appendix~\ref{appendix:proofs}. In \citet{dpo}, two loss functions are proposed under two ranking frameworks: Bradley-Terry (BT) model~\citep{bradley1952rank} and Plackett-Luce (PL) model~\citep{plackett1975analysis}. We refer $\text{DPO}_\text{BT}$ loss as the BT model-based pairwise loss, and refer $\text{DPO}_\text{PL}$ loss as the PL model-based listwise loss. We use $\text{SLiC}_\text{norm}$ to denote the improved SLiC~\citep{slic} as in RSO~\citep{rso}, which normalizes the likelihood with the reference policy and removes the regularization.
\paragraph{Pairwise preference losses.} The pairwise logistic ranking loss~\citep{burges2005learning} is one popular choice to fit a list of ranked data: $\mathcal{L}_\text{pair-logistic}(\pi_{\theta};\pi_\textrm{ref}, \beta) =$
\begin{equation}
\label{eq:ranknet}
   \resizebox{0.8\hsize}{!}{$    \mathbb{E}_{x, \mathbf{y}, \mathbf{\psi} \sim \mathcal{D}}  \left[ \sum_{\psi_i>\psi_j}\log(1+e^{-(s_i-s_j)}) \right].$}
\end{equation}
We connect the above loss with $\text{DPO}_\text{BT}$ via the following proposition:
\begin{proposition}\label{prop:dpo}
%When $K=2$, Bradley-Terry (BT) model is used as the label likelihood function $P(\psi|\mathbf{s},\tau)$, and permutation likelihood function $P(\tau | \mathbf{s})$ is a delta distribution with probability mass $=1$ only on the label induced permutation, LiPO is equivalent to DPO~\citep{dpo}.
%When $K=2$, Bradley-Terry (BT) model is used as the label likelihood function $P(\psi|\mathbf{s},\tau)$, and permutation likelihood function $P(\tau | \mathbf{s})$ is a delta distribution with probability mass only on the label induced permutation, LiPO is equivalent to DPO~\citep{dpo}.
When $K=2$ and pairwise logistic ranking loss is used, LiPO is equivalent to $\text{DPO}_\text{BT}$~\citep{dpo}.
\end{proposition}

Similarly, we can connect $\text{SLiC}_\text{norm}$ with pairwise hinge loss from RankSVM~\citep{joachims2002optimizing}: $ \mathcal{L}_{\text{pair-hinge}}(\pi_{\theta};\pi_\textrm{ref},\beta) =$
% \vspace{-0.5em}
\begin{equation}
\label{eq:ranksvm}
      \resizebox{0.8\hsize}{!}{$  \mathbb{E}_{x, \mathbf{y}, \mathbf{\psi} \sim \mathcal{D}}  \left[\sum_{\psi_i>\psi_j}\max\left(0, 1 - (s_i-s_j)\right)\right]$}
\end{equation}
\begin{proposition}\label{prop:slic}
When $K=2$ and pairwise hinge ranking loss is used, LiPO is equivalent to $\text{SLiC}_\text{norm}$~\citep{slic,rso}.
\end{proposition}

\paragraph{Listwise preference losses.} One can fit all pairs using pairwise-logistic or pairwise-hinge losses. Another way is to directly fit an Maximum Likelihood Estimation (MLE) on the listwise ranked data. \citet{xia2008listwise} proposes list MLE ranking loss: $\mathcal{L}_{\text{list-mle}}(\pi_{\theta};\pi_\textrm{ref},\beta) =$
\begin{equation}\label{eq:list-mle}
 \resizebox{0.9\hsize}{!}{$ -\mathbb{E}_{x, y_1, y_2... y_K\sim \mathcal{D}} \left[\log \prod_{k=1}^{K} \frac{\exp(s_{\tau(k)})}{\sum_{j=k}^{K}\exp(s_{\tau(j)})}\right],$}
\end{equation}
where $\tau(i)$ is the document ranked at the $i$-th position in the listwise permutation determined by label. This loss is used in $\text{DPO}_\text{PL}$~\citep{dpo} and PRO~\citep{song2023preference}.
\begin{proposition}\label{prop:mle}
When the list MLE loss is used, LiPO is equivalent to $\text{DPO}_\text{PL}$~\citep{dpo} and PRO~\citep{song2023preference}.
\end{proposition}

% We note that the DPO paper only briefly discussed the $\text{DPO}_\text{PL}$ in their appendix without any thorough study. Though a straightforward extension from the above pairwise model, however, we show that $\text{DPO}_\text{PL}$ is not very competitive as a ranking objective in experiments. 
%\paragraph{Discussion.} 

\subsection{Pitfalls of Existing Work}

From the general LiPO formulation and the analysis of recent policy optimization methods, we can see they map to specific choices of existing ranking objectives. Through LTR, we note there are two major concerns of the discussed methods that may limit the effectiveness of preference optimization. 

First, all pairwise approaches, which dominate the current preference optimization literature, ignore listwise permutation information beyond pairs. Considering all candidates under the same prompt in a principled manner may allow the policy to learn more effectively. 

Second, virtually all existing methods, even the listwise one ($\text{DPO}_\text{PL}$ and PRO), ignore the label values,
% IPO does not?
i.e., they only focus on the optimal pairwise or listwise \textit{rank-ordering} of the responses. This has information loss and may incur confusing learning behavior. For example, two lists with labels (0.99, 0.50, 0.01) and (0.51, 0.50, 0.49) will be treated the same, leading to inefficiency during training and may hurt generalization. They also force an ordering while it is common to have tied labels in ranking data~\citep{8186875}. 
\section{LiPO-$\lambda$}\label{sec:lipo-lambda}
Under the general LiPO framework, a rich family of other optimization objectives can be explored from the LTR literature. In this section, we propose a specific instantiation, LiPO-$\lambda$, which builds upon a state-of-the-art ranking objective, addresses the above pitfalls  in a principled manner, and performs well empirically in the experiments.

Our LiPO-$\lambda$ is based on the LambdaLoss method~\citep{NIPS2006_af44c4c5,wang2018lambdaloss}. Specifically, the training objective of LiPO-$\lambda$ is:
$\mathcal{L}_{\text{lambda-loss}}(\pi_{\theta};\pi_\textrm{ref},\beta)= $
\begin{equation}\label{eq:lipo-lambda}
   \resizebox{0.8\hsize}{!}{$ \mathbb{E}_{x, \mathbf{y}, \mathbf{\psi} \sim \mathcal{D}} \left[ \sum_{\psi_i>\psi_j} \Delta_{i,j} \log (1+e^{-(s_i-s_j)})\right],$}
\end{equation}
where $\Delta_{i,j} =  |G_i-G_j|\cdot|\frac{1}{D(\tau(i))}-\frac{1}{D(\tau(j))}|$, known as the Lambda weight. $G$ is called a gain function with $G_i = 2^{\psi_i} - 1$ as the commonly used one. $D$ is a rank discount function with $D(\tau(i)) = \log (1 + \tau(i))$ as the commonly used one, where $\tau(i)$ is the rank position of $y_i$ in the ranking permutation induced by $\mathbf{s}$, thus it is a \textit{listwise} method even though the formula can be written in terms of pairs. In other words, there are dependencies on other items in the same list for each pair. One has the flexibility to change the gain and discount functions and we resort to the original LambdaLoss configuration for its simplicity and strong empirical performance. It has been shown that this loss function can optimize the DCG metric~\citep{NIPS2006_af44c4c5,10.1145/1571941.1572021},
\begin{equation*}
   \resizebox{0.4\hsize}{!}{ $DCG = \sum_{i=1}^K \frac{G_i}{D(\tau(i))},$}
\end{equation*}
which has several ideal properties as a well-founded ranking metric such as consistent distinguishability~\citep{wang2013theoretical}: for every pair of substantially different ranking policies, the ranking
metric can decide which one is better in a consistent manner on almost all datasets. %Linking back to 

\paragraph{Comparisons.} There are several interesting findings by comparing LiPO-$\lambda$ with other methods under the LiPO framework. First, the gain function $G$ considers label score values $\psi$, which is ignored in virtually all existing methods. Second, comparing with the list MLE loss, the permutation considered is induced by the model prediction scores, instead of the static labels. \citet{NIPS2006_af44c4c5} showed that considering such dynamic permutations based on model predictions during training can lead to smoother optimization landscapes to optimize the non-smooth ranking objectives, resulting in better empirical performance than using the static ranks from labels. Last but not least, LiPO-$\lambda$ can be treated as a weighted version of $\text{DPO}_\text{BT}$ over all pairs of ranked list by comparing with Eq~\ref{eq:ranknet}. Instead of treating each pair equally, Lambda weight is a listwise permutation aware weighting mechanism. One intuition of Lambda weight is to weight each pairs by the difference of ranking metrics when they are swapped in the list~\citep{burges2010ranknet}. 

\section{Other Ranking Losses}
Other existing or future ranking objectives may be studied under the LiPO framework. In experiments, in addition to existing methods that can be treated as specific instantiations of LiPO, we also study two pointwise ranking objectives, the pointwise Mean Squared Error (MSE) loss: $\mathcal{L}_{\text{point-mse}}(\pi_{\theta};\pi_\textrm{ref},\beta) =$
\begin{equation}\label{eq:pointwise-mse}
 \resizebox{0.5\hsize}{!}{$ \mathbb{E}_{x, \mathbf{y}, \mathbf{\psi} \sim \mathcal{D}}\sum_{k=1}^K(\psi_k-s_k)^2,$}
\end{equation}
and the pointwise sigmoid cross entropy loss: $\mathcal{L}_{\text{point-sigmoid}}(\pi_{\theta};\pi_\textrm{ref},\beta) =$
\begin{equation}\label{eq:pointwise-sigmoid}
\resizebox{1.01\hsize}{!}{$ -\mathbb{E}_{x, \mathbf{y}, \mathbf{\psi} \sim \mathcal{D}}\sum_{i=1}^K\left(\psi_i\log\sigma(s_i) + (1-\psi_i)\log(1-\sigma(s_i))\right)$}
\end{equation}

We also consider softmax cross entropy loss as in ListNet~\citep{cao2007learning}), which again maps to a recent work called NCE~\cite{chen2024noise}:$\mathcal{L}_{\textrm{softmax}}(\pi_{\theta};\pi_\textrm{ref},\beta) =$
\begin{equation}
\label{eq:softmax}
 \resizebox{0.9\hsize}{!}{$     \mathbb{E}_{x, \mathbf{y}, \mathbf{\psi} \sim \mathcal{D}}  \left[\sum_{k=1}^{K} \frac{\psi_k}{\sum_{j=1}^{K}\psi_j} \log\left( \frac{\exp(s_k)}{\sum_{j=1}^{K}\exp(s_j)} \right) \right].$}
\end{equation}

\section{Experiments}\label{sec:exp}

\paragraph{Tasks.} We study different ranking losses unified under the LiPO framework on the popular Reddit TL;DR summarization~\citep{stiennon2020learning} and AnthropicHH
dialogue~\citep{bai2022training} datasets, and provide further verification on OpenAssistant dataset in Section~\ref{human_ranked_data}. The Reddit TL;DR summarization dataset contains both fine-tuning data $\mathcal{D}^{\textrm{tldr}}_{\textrm{sft}}$ and human feedback data $\mathcal{D}^{\textrm{tldr}}_{\textrm{hf}}$. $\mathcal{D}^{\textrm{tldr}}_{\textrm{sft}}$ contains 117k/6k examples in train and validation splits. $\mathcal{D}^{\textrm{tldr}}_{\textrm{hf}}$ consists of 93k human preferences on decodes from multiple models.
The AnthropicHH is a dialogue dataset with $x$ as conversation between a human query and an AI assistant. We use the helpful slice $\mathcal{D}^{\textrm{helpful}}_{\textrm{hf}}$ from 161k/9k examples in train and validation splits. We use the positive responses as SFT targets.

\paragraph{Method.} For each task, we first train a T5-large (770M)~\citep{raffel2020exploring} SFT policy on the SFT dataset. We pick the best checkpoint with lowest perplexity on validation split. We also train a T5-XXL (11B) pairwise reward-ranking model~\citep{slic,rso} on the human preference dataset (See Appendix~\ref{appendix:reward-ranking model} for details). We pick the best checkpoint with the highest accuracy on validation split of the human preference dataset. Then we sample $K=8$ responses (with ablation below) for each prompt from the SFT policy via Top-K sampling with $temperature = 0.7$ and $top\_k = 40$. We conduct all pair comparisons using the pairwise reward-ranking model, resulting in a winning probability matrix $\Psi\in [0,1]^{K\times K}$, after which we compute $\psi_k$ as $\psi_k = \frac{1}{K}\sum_{i=1}^K\Psi_{ki}$. Reward model is shown to be critical in the success of alignment~\citep{rso,gao2023scaling}. This aggregation and labeling schema is valid theoretically~\citep{shah2018simple} and works well empirically~\citep{qin2023large} in the ranking literature. For directly optimizing on human-ranked responses without a reward model, we approximate the winning probability as $\hat{\mathbb{P}}(y_i\succ y_j)=\mathbf{1}(\text{rank $y_i$ is higher than $y_j$})$. See Section~\ref{human_ranked_data} for an experiment on OpenAssisant dataset~\citep{kopf2024openassistant}.
To compute the normalized ranking scores $\mathbf{s}$, we set $\beta=0.05$. 

We consider three types of loss functions under the LiPO framework: pointwise, pairwise, and listwise. Pointwise losses include point-mse (Eq~\ref{eq:pointwise-mse}) and point-sigmoid (Eq~\ref{eq:pointwise-sigmoid}). Pairwise losses include pair-hinge (Eq~\ref{eq:ranksvm}) and  pair-logistic (Eq~\ref{eq:ranknet}). Listwise losses include list-mle (Eq~\ref{eq:list-mle}), softmax (Eq~\ref{eq:softmax}), and 
lambda-loss (Eq~\ref{eq:lipo-lambda}). We use the open-sourced RAX library~\citep{jagerman2022rax} to compute losses in Jax. See Appendix \ref{appendix:example_usage} for example usages.

For pairwise losses, we utilize all 64 pairs and show in Section~\ref{sec:ablations} that it can \textit{improve} performance than sampling a single pair each time as commonly done, so this setting is not only fair but also benefit pairwise methods. We use batch size 32 and learning rate 2e-5 with Adafactor optimizer~\citep{shazeer18a}, which takes about 1 day to run the calibration on 32 TPU-v3 chips. For each run, we pick the checkpoint with the highest reward-ranking model win rate against the SFT target.

\paragraph{Evaluation.} Our experiments use three different approaches to evaluate following existing protocols~\citep{dpo,rso}: Proxy Reward Model, AutoSxS, and human evaluation. Proxy Reward Model computes the relative win rate
of generated response against SFT target over the SFT validation split on the trained T5-XXL pairwise reward-ranking model. AutoSxS computes the relative win rate
of generated response against SFT target over the SFT validation split using large instruction tuned model (PaLM 2-L-IT)~\citep{palm2} via few-shot in-context learning (details in Appendix~\ref{appendix:autosxs}). Human Evaluation asks human raters to assign an absolute quality score on each response and determine the relatively best one among $\text{DPO}_\text{BT}$, $\text{DPO}_\text{PL}$ (PRO) and LiPO-$\lambda$ (details in Appendix~\ref{appendix:human_sxs}). We note these are standard evaluation protocols in related work.

\begin{table*}[ht]\small
\begin{center}
\begin{tabular}{llcc}
\hline
 Approach &  Ranking Loss & Proxy Reward (\%) & AutoSxS (\%) 
\\ 
\hline\hline
\multicolumn{4}{c}{\textbf{Reddit TL;DR}}\\
\hline
   & point-mse & $49.43_{\substack{\pm 1.18}}$ & $39.94_{\substack{\pm 1.22}}$ \\
   & point-sigmoid & $64.14_{\substack{\pm 1.16}}$ & $49.28_{\substack{\pm 1.27}}$ \\
 NCE   & softmax & $75.40_{\substack{\pm 0.98}}$ & $58.60_{\substack{\pm 1.22}}$ \\  
 $\text{SLiC}_\text{norm}$ & pair-hinge & $87.23_{\substack{\pm 0.78}}$ & $67.16_{\substack{\pm 1.15}}$ \\ 
 $\text{DPO}_\text{BT}$  & pair-logistic & $88.52_{\substack{\pm 0.74}}$ & $67.09_{\substack{\pm 1.17}}$ \\ 
 $\text{DPO}_\text{PL}$ (PRO) & list-mle & $88.27_{\substack{\pm 0.76}}$ & $67.13_{\substack{\pm 1.08}}$ \\
 LiPO-$\lambda$ & lambda-loss & $\textbf{90.60}_{\substack{\pm 0.65}}$ & $\textbf{68.26}_{\substack{\pm 1.05}}$ \\
\hline
\hline
\multicolumn{4}{c}{\textbf{AnthropicHH}}\\
\hline
   & point-mse & $57.55_{\substack{\pm 1.22}}$ & $21.97_{\substack{\pm 0.85}}$ \\
   & point-sigmoid & $71.35_{\substack{\pm 1.11}}$ & $25.72_{\substack{\pm 0.87}}$ \\
  NCE & softmax & $73.21_{\substack{\pm 1.07}}$ & $28.87_{\substack{\pm 0.95}}$ \\  
 $\text{SLiC}_\text{norm}$ & pair-hinge & $89.68_{\substack{\pm 0.72}}$ & $42.07_{\substack{\pm 1.01}}$ \\ 
 $\text{DPO}_\text{BT}$  & pair-logistic & $91.11_{\substack{\pm 0.66}}$ & $44.80_{\substack{\pm 1.00}}$ \\ 
 $\text{DPO}_\text{PL}$ (PRO) & list-mle & $90.61_{\substack{\pm 0.72}}$ & $43.25_{\substack{\pm 1.02}}$ \\
 LiPO-$\lambda$ & lambda-loss & $\textbf{92.60}_{\substack{\pm 0.62}}$ & $\textbf{47.90}_{\substack{\pm 0.98}}$ \\
\hline
\end{tabular}
\caption{Comparison of different methods with T5-large policy model to leverage listwise preference data. Proxy rewards and few-shot PaLM 2-L-IT win rates against SFT target text are reported. All methods use preference list with size 8, and pairwise methods including $\text{SLiC}_\text{norm}$ and $\text{DPO}_\text{BT}$ use all pairs generated from the list and treat them equally. 95\% bootstrapped confidence intervals are indicated by the subscripts.}
%https://docs.google.com/spreadsheets/d/1dWKHtbT3ipoU-qTF15dYEQpoQG2u0pRP63XfOBOSsKM/edit?resourcekey=0-9om9VIsSoTkiIEucJD5vlQ#gid=0
\label{tab:main_results}
\end{center}
\end{table*}

\subsection{Performance Comparison}
The main comparison results are shown in Table~\ref{tab:main_results}. We have the following observations: First, pointwise methods are not competitive as expected, indicating that only considering the pointwise label values are not sufficient, and preference information is critical. Second, $\text{DPO}_\text{PL}$ (PRO) does not perform better than $\text{DPO}_\text{BT}$, showing that the choice of listwise ranking objective is important given listwise data. This aligns with existing LTR literature that list-mle is not a competitive ranking objective as it enforces listwise permutation without caring about label values. On the other hand, while $\text{DPO}_\text{BT}$ does not consider label values either, the pairwise format can be less prone to ill-behaved listwise behaviors. Third, the listwise Softmax loss (NCE) is not competitive. This is understandable as Softmax loss is most effective on optimizing listwise ranking with sparse labels, such as binary click data~\citep{6867349}. For LM generation, the responses are sampled from a plausible set so the labels are dense, which do not fit the Softmax loss well. Meanwhile, LiPO-$\lambda$ shows strong performance and improves upon all baselines by effectively leveraging listwise data and label value information. 
\subsection{Ablation Studies and Analysis}
\label{sec:ablations}
\begin{figure*}
\centering
% https://colab.corp.google.com/drive/1yyDOgLn6goobILuukMb9c2tOoLywckDf?usp=sharing
\begin{subfigure}[b]{0.40\textwidth}
\centering
\includegraphics[width=\textwidth]{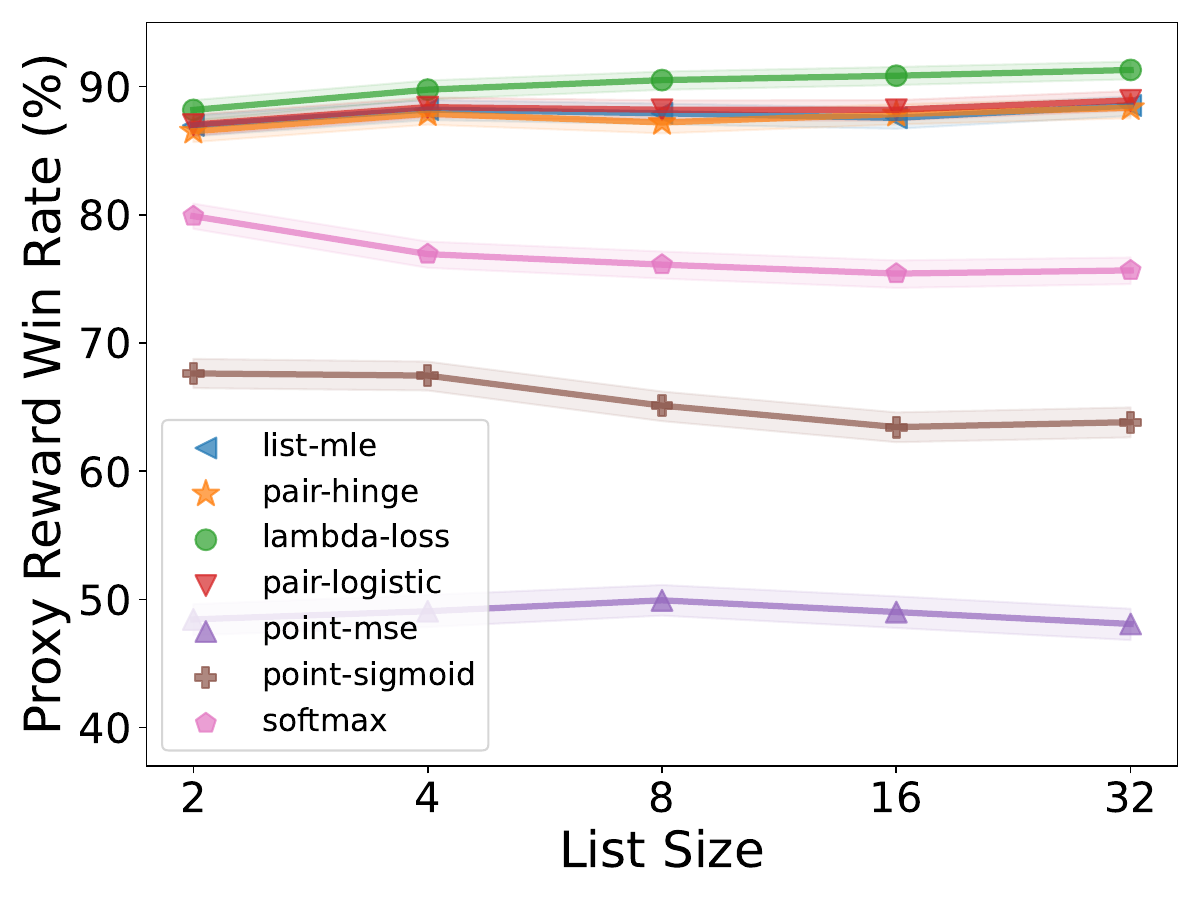}
\caption{}
%   \caption{Performance of different ranking losses with varying list sizes on the Reddit TL;DR dataset. The shaded areas are 95\% confidence intervals. 95\% bootstrapped confidence intervals are shown as shaded areas.}
% \label{fig:listsize}
\end{subfigure}
\quad\quad
% \hspace{1em}
% https://colab.corp.google.com/drive/1yyDOgLn6goobILuukMb9c2tOoLywckDf?usp=sharing
\begin{subfigure}[b]{0.37\textwidth}
\centering
\includegraphics[width=\textwidth]{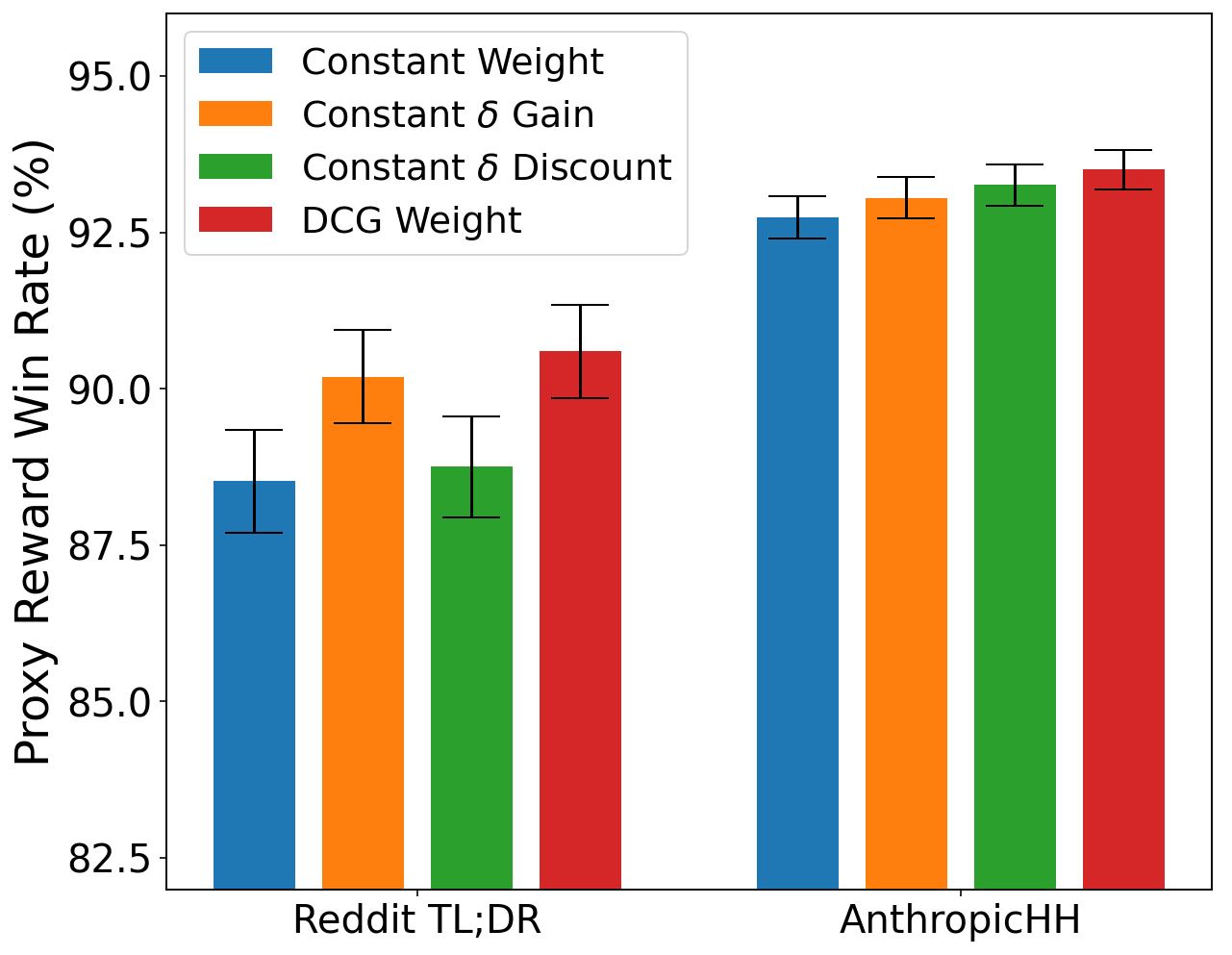}
\caption{}
%   \caption{Performance on the Reddit TL;DR and AnthropicHH datasets by using different Lambda weight choices. See text for explanation of different options. The error bars denote 95\% confidence intervals.}
% \label{fig:lambdaweight}
\end{subfigure}
\caption{\textbf{(a):} Performance of different ranking losses with varying list sizes on the Reddit TL;DR dataset. The shaded areas are 95\% confidence intervals. 95\% bootstrapped confidence intervals are shown as shaded areas. \textbf{(b):} Performance on the Reddit TL;DR and AnthropicHH datasets by using different Lambda weight choices. See text for explanation of different options. The error bars denote 95\% confidence intervals.}
% \vspace{-1.5em}
\label{fig:listsize_lambdaweight}
\end{figure*}

To gain an in-depth understanding of the benefits brought by the listwise formulation and methods, we conduct ablation studies to understand the effect in terms of listwise data, Lambda weight choices, and model sizes. 
\paragraph{Ablation study on list size.}
To better understand the effect of preference optimization on listwise data, we conduct analysis over multiple choices of list sizes on the Reddit TL;DR dataset. As illustrated in Figure~\ref{fig:listsize_lambdaweight}(a), most methods can benefit from going beyond pairwise training data (List Size = 2) by leveraging more information under each prompt $x$. LiPO-${\lambda}$ with lambda-loss is the only method that can robustly benefit from longer list sizes, showing it can more effectively leverage the rich listwise information.

\paragraph{Ablation study on Lambda weights.}
As discussed in Section~\ref{sec:lipo-lambda}, we use specific choice of Lambda weights by setting the gain function $G_i = 2^{\psi_i} - 1$ and rank discount function $D(\tau(i)) = \log (1 + \tau(i))$, which is called the DCG weight, that can optimize the DCG ranking metric. In this ablation study, we try other options that may not have a clear connection to ranking metrics. Constant Weight assigns equal weights on all pairs, which reduces to DPO$_{\textrm{BT}}$. Constant $\delta$ Gain sets the gain function difference to a constant: $|G_i-G_j| = 1$, and Constant $\delta$ Discount sets the rank discount function difference to a constant: $|\frac{1}{D(\tau(i))}-\frac{1}{D(\tau(j))}| = 1$. The comparisons in Figure~\ref{fig:listsize_lambdaweight}(b) show that using DCG weight is most effective on both datasets, showing the importance of setting both functions appropriately in order to optimize well-founded ranking metrics.

\begin{table}[!htb]
\begin{minipage}{0.9\linewidth}
\resizebox{\columnwidth}{!}{\begin{tabular}{lcc}
\hline
Approach  & Proxy Reward (\%) & AutoSxS (\%) 
\\ 
\hline\hline
\multicolumn{3}{c}{\textbf{Reddit TL;DR}}\\
\hline
$\text{DPO}_\text{BT}$ & $96.22_{\substack{\pm 0.43}}$  &  $82.38_{\substack{\pm 0.85}}$\\
$\text{DPO}_\text{PL}$ (PRO) & $96.00_{\substack{\pm 0.45}}$  &  $81.96_{\substack{\pm 0.92}}$\\
LiPO-$\lambda$ & $\textbf{97.32}_{\substack{\pm 0.36}}$  & $\textbf{83.79}_{\substack{\pm 0.88}}$ \\\hline
\hline
\multicolumn{3}{c}{\textbf{AnthropicHH}}\\
\hline
$\text{DPO}_\text{BT}$ &  $97.48_{\substack{\pm 0.34}}$ &  $68.81_{\substack{\pm 0.94}}$\\
$\text{DPO}_\text{PL}$ (PRO) &  $97.28_{\substack{\pm 0.37}}$ &  $68.84_{\substack{\pm 0.91}}$\\
LiPO-$\lambda$ & $\textbf{98.27}_{\substack{\pm 0.29}}$  & $\textbf{69.81}_{\substack{\pm 0.91}}$ \\\hline
\end{tabular}}
\caption{Comparison of $\text{DPO}_\text{BT}$, $\text{DPO}_\text{PL}$ (PRO), and LiPO-$\lambda$ with T5-XXL policy model. Proxy rewards and AutoSxS win rates against SFT target text are reported. All methods use preference list with size 8. 95\% bootstrapped confidence intervals are indicated by the subscripts.}
\label{tab:xxl_results}
\end{minipage}\hfill

\begin{minipage}{.9\linewidth}
\resizebox{\columnwidth}{!}{\begin{tabular}{lcc}
\hline
 Approach  &  Chosen as Preferred & Quality
\\ 
\hline\hline
\multicolumn{3}{c}{\textbf{Reddit TL;DR}}\\
\hline
$\text{DPO}_\text{BT}$ & 19\% & 3.63 \\
$\text{DPO}_\text{PL}$ (PRO) & 16\% & 3.67 \\
LiPO-$\lambda$&\textbf{40\%} & \textbf{3.80}  \\
\hline\hline
\multicolumn{3}{c}{\textbf{AnthropicHH}}\\
\hline
$\text{DPO}_\text{BT}$ & 20\% & 3.66 \\
$\text{DPO}_\text{PL}$ (PRO) & 20\% & 3.66 \\
LiPO-$\lambda$&\textbf{27\%} & \textbf{3.72}  \\
\hline
\end{tabular}}
\caption{Human evaluation comparing three loss functions on two tasks. The proportion may not sum up to 100\% due to equal preference cases.}
\label{tab:human_eval_result}
\end{minipage}
\end{table}

\paragraph{Scale up the policy model.}
To understand how well the LiPO can be scaled up to larger policy models, we train a T5-XXL policy model and compare among $\text{DPO}_\text{BT}$, $\text{DPO}_\text{PL}$ (PRO) and LiPO-$\lambda$. Table~\ref{tab:xxl_results} shows that all three methods scale up well and LiPO-$\lambda$ is competitive on both tasks.\footnote{We further conduct human evaluation study to verify the gains in Appendix~\ref{appendix:human_eval_xxl}.}

% \paragraph{Likelihood analysis}
% To better understand the effect of Lambda weight in LiPO-$\lambda$ loss, we compare the the rank of $\psi$ versus the rank of $s$. TBD

\subsection{Human Evaluation Results}\label{sec:human_sxs}

To further verify the improvements of LiPO-$\lambda$, we conduct side-by-side human evaluation using Amazon Mechanical Turk. Given a document and three responses generated from $\text{DPO}_\text{BT}$, $\text{DPO}_\text{PL}$ (PRO) and LiPO-$\lambda$, raters are asked to assign a pointwise overall quality (1-5) to each response, and to choose the best one. 
Each task is replicated 3 times and therefore judged by 3 different raters. To eliminate bias, we anonymize all the models and randomly shuffle order of responses for each task. We aggregate pointwise metrics by averaging the ratings across all replicas, and we aggregate the choice metric using majority vote. For more details about the tasks, see Appendix~\ref{appendix:human_sxs}.

In total 50 different raters participated in the Reddit TL;DR evaluation study with a median of 24.5 tasks per rater. The human evaluation results are shown in Table~\ref{tab:human_eval_result}. LiPO-$\lambda$ has shown to be better than $\text{DPO}_\text{BT}$ and $\text{DPO}_\text{PL}$ (PRO) in both tasks.

\subsection{Direct Alignment on Human Ranked Responses}
\label{human_ranked_data}
To further verify the case when only ranks are available, we add a new dataset, OpenAssistant~\citep{kopf2024openassistant}. The dataset is of conversation tree structure with only rank associated with each response. We filter out non-English prompts and leave the prompts with exactly 3 ranked responses. We get 2.6k and 155 prompts in training and validation splits, respectively. We pick the best response as the target during SFT, followed by listwise calibration. To compute the label, we first construct the preference matrix $M$ as $M_{ii}=0.5$ and $M_{ij}=\mathbf{1}(\text{rank $y_i$ is higher than $y_j$})$
 (see Section~\ref{sec:formulation}). Then we aggregate the matrix as label values according to Section~\ref{sec:exp}. The results are summarized in Table~\ref{tab:oasst}. LiPO-$\lambda$ again performs the best even with rank-only annotations directly from human.

\begin{table}[h]\small
\begin{center}
\begin{tabular}{lcc}
\hline
 Approach  &  Ranking Loss & AutoSxS (\%)
\\ 
\hline
$\text{DPO}_\text{BT}$ & pair-logistic & 25.48 \\
$\text{DPO}_\text{PL}$ (PRO) & list-mle & 23.54 \\
LiPO-$\lambda$& lambda-loss & \textbf{27.10}  \\
\hline
\end{tabular}
\caption{Comparison of $\text{DPO}_\text{BT}$, $\text{DPO}_\text{PL}$ (PRO), and LiPO-$\lambda$ with T5-large policy model on OpenAssistant dataset. Few-shot PaLM 2-L-IT win rates against SFT target text are reported. All methods use preference list with size 3.}
\label{tab:oasst}
\end{center}
\end{table}

%\section{Discussions}
%\paragraph{Generative Ranking.} We treat LM alignment as a listwise ranking problem, the problem of generative ranking of potential responses and draw insights from the IR field. Though traditional IR methods score each query $x$ against each candidate document explicitly, recent IR efforts~\citep{tay2022transformer,wang2022neural} start to explore generative retrieval / ranking with the Transformer architecture, where only $x$ is provided during inference and document identifiers (which can be several tokens) are directly generated. This is more similar to the LM generation scenario and provides more evidence to our ranking perspective of the problem. 
\section{Related Work}
\paragraph{LM Alignment.} While self-supervised LMs learn to complete some interesting tasks~\citep{radford2019language}, their performance on downstream tasks, such as acting as a conversational agent, can be significantly improved by alignment with human preference datasets. The pivotal Reinforcement Learning from Human Feedback (RLHF) framework~\citep{rlhf} first fits a reward function under a preference model such as the Bradley-Terry model~\citep{bradley1952rank}, then fine-tunes the LM to maximize the given reward using reinforcement learning algorithms. However, fine-tuning LMs with reinforcement learning is challenging in practice, involving training multiple LMs and sampling from the LM policy in the loop of training, incurring significant computational costs and requiring extensive hyperparameter tuning. 

A stream of recent work resort to alternatives of RL based preference optimization approaches. As we discussed, two parallel work, SLiC~\citep{slic} and RRHF~\citep{rrhf} directly use human preference data or use a reward model to label preferences, then both use a pairwise hinge loss to align policy responses. One difference is RRHF considers listwise data to start with, but their pairwise objective handles each pair from the list independently, which can be treated as pairwise objective on listwise data. DPO~\citep{dpo} proposes to directly align the behavior of LM without using a reward model with the pairwise logistic loss. IPO~\citep{ipo,calandriello2024human} improves DPO by removing the Bradley-Terry assumption. RSO~\citep{rso} improves DPO with a better sourcing of preference dataset. GSHF~\citep{xiong2023gibbs,dong2024rlhf} further improves RSO with iterative optimization. OAIF~\citep{guo2024direct} improves DPO with online AI feedback. Same as listwise version of DPO, PRO~\citep{song2023preference} proposes list MLE loss for list of responses. In this work, we mainly focus on the optimization objective perspective of LM alignment.

\paragraph{Learning-to-Rank.} 
The Learning-to-Rank (LTR) field has a rich literature due to its practical values in applications such as web search~\citep{8186875}. Traditional LTR work mainly focuses on developing more effective ranking objectives to optimize ranking metrics, ranging from pointwise, pairwise, to listwise approaches~\citep{8186875}. Beside ranking objectives, the LTR is also concerned about other relevant topics such as ranking architectures~\cite{qin2021neural} and learning from biased feedback~\cite{wang2021non}, which are orthogonal to our work since we focus on standard LLM architectures and assume reliable feedback from reward models.

RankSVM~\citep{joachims2002optimizing} and RankNet~\citep{burges2005learning} leverage pairwise hinge loss and pairwise logistic loss respectively for the ranking problem. Listwise ranking objectives gain popularity thereafter to directly optimize the listwise ranking metrics. ListMLE and Softmax cross entropy losses are two representative listwise losses proposed in~\citep{xia2008listwise} and ~\citep{cao2007learning}. ListMLE only concerns about the ordering under the Plackett-Luce model~\citep{luce2005individual}, and Softmax cross entropy loss is effective on lists with sparse labels, such as click logs~\citep{6867349,bai2023regression}. LambdaRank~\citep{NIPS2006_af44c4c5} shows that weighting pairs with the listwise Lambda weight leads to strong empirical performance in terms of optimizing the non-smooth DCG metric, and it is unified under the LambdaLoss~\citep{wang2018lambdaloss} framework with theoretical justification and convergence proof. In this work, we show the connection between existing LM alignment methods and ranking objectives from LTR. Following this connection, we studied multiple under-explored listwise objectives for LM alignment.

\section{Conclusion}
We describe the LiPO framework for LM alignment with a list of responses for each prompt with connection to LTR techniques. We generalize recent preference optimization methods and analyze pitfalls of existing methods from the LTR perspective. With comprehensive studies over existing LTR losses, we highlight LiPO-$\lambda$ as the best approach that builds upon the state-of-the-art ranking objectives and shows competitive performance across several tasks.
%and (3) We can study LiPO on other large models such as Gemini to further understand its scaling effects.

\section{Limitations}\label{sec:limitations}
In this work, we study the algorithms on listwise responses only in offline setting, same as pivot work in the literature such as DPO. It is interesting to study how to do the online learning where the list of responses are elicited from the policy being trained to reduce the distribution shift. Also, the number of labels generated by the reward model is quadratic to list size. Even though in the offline setting, the cost is only realized in the preprocessing step, for future work, we may leverage more effective approaches, such as using partial comparisons to reconstruct the label relevance score.

\bibliography{custom}

\begin{thebibliography}{48}
\providecommand{\natexlab}[1]{#1}

\bibitem[{Azar et~al.(2023)Azar, Rowland, Piot, Guo, Calandriello, Valko, and
  Munos}]{ipo}
Mohammad~Gheshlaghi Azar, Mark Rowland, Bilal Piot, Daniel Guo, Daniele
  Calandriello, Michal Valko, and R{\'e}mi Munos. 2023.
\newblock A general theoretical paradigm to understand learning from human
  preferences.
\newblock \emph{arXiv preprint arXiv:2310.12036}.

\bibitem[{Bai et~al.(2023)Bai, Jagerman, Qin, Yan, Kar, Lin, Wang, Bendersky,
  and Najork}]{bai2023regression}
Aijun Bai, Rolf Jagerman, Zhen Qin, Le~Yan, Pratyush Kar, Bing-Rong Lin,
  Xuanhui Wang, Michael Bendersky, and Marc Najork. 2023.
\newblock Regression compatible listwise objectives for calibrated ranking with
  binary relevance.
\newblock In \emph{Proceedings of the 32nd ACM International Conference on
  Information and Knowledge Management}, pages 4502--4508.

\bibitem[{Bai et~al.(2022)Bai, Jones, Ndousse, Askell, Chen, DasSarma, Drain,
  Fort, Ganguli, Henighan et~al.}]{bai2022training}
Yuntao Bai, Andy Jones, Kamal Ndousse, Amanda Askell, Anna Chen, Nova DasSarma,
  Dawn Drain, Stanislav Fort, Deep Ganguli, Tom Henighan, et~al. 2022.
\newblock Training a helpful and harmless assistant with reinforcement learning
  from human feedback.
\newblock \emph{arXiv preprint arXiv:2204.05862}.

\bibitem[{Bradley and Terry(1952)}]{bradley1952rank}
Ralph~Allan Bradley and Milton~E Terry. 1952.
\newblock Rank analysis of incomplete block designs: I. the method of paired
  comparisons.
\newblock \emph{Biometrika}, 39(3/4):324--345.

\bibitem[{Burges et~al.(2005)Burges, Shaked, Renshaw, Lazier, Deeds, Hamilton,
  and Hullender}]{burges2005learning}
Chris Burges, Tal Shaked, Erin Renshaw, Ari Lazier, Matt Deeds, Nicole
  Hamilton, and Greg Hullender. 2005.
\newblock Learning to rank using gradient descent.
\newblock In \emph{Proceedings of the 22nd international conference on Machine
  learning}, pages 89--96.

\bibitem[{Burges et~al.(2006)Burges, Ragno, and Le}]{NIPS2006_af44c4c5}
Christopher Burges, Robert Ragno, and Quoc Le. 2006.
\newblock Learning to rank with nonsmooth cost functions.
\newblock In \emph{Advances in Neural Information Processing Systems},
  volume~19. MIT Press.

\bibitem[{Burges(2010)}]{burges2010ranknet}
Christopher~JC Burges. 2010.
\newblock From ranknet to lambdarank to lambdamart: An overview.
\newblock \emph{Learning}, 11(23-581):81.

\bibitem[{Calandriello et~al.(2024)Calandriello, Guo, Munos, Rowland, Tang,
  Pires, Richemond, Lan, Valko, Liu et~al.}]{calandriello2024human}
Daniele Calandriello, Daniel Guo, Remi Munos, Mark Rowland, Yunhao Tang,
  Bernardo~Avila Pires, Pierre~Harvey Richemond, Charline~Le Lan, Michal Valko,
  Tianqi Liu, et~al. 2024.
\newblock Human alignment of large language models through online preference
  optimisation.
\newblock \emph{arXiv preprint arXiv:2403.08635}.

\bibitem[{Cao et~al.(2007)Cao, Qin, Liu, Tsai, and Li}]{cao2007learning}
Zhe Cao, Tao Qin, Tie-Yan Liu, Ming-Feng Tsai, and Hang Li. 2007.
\newblock Learning to rank: from pairwise approach to listwise approach.
\newblock In \emph{Proceedings of the 24th international conference on Machine
  learning}, pages 129--136.

\bibitem[{Chen et~al.(2024)Chen, He, Su, and Zhu}]{chen2024noise}
Huayu Chen, Guande He, Hang Su, and Jun Zhu. 2024.
\newblock Noise contrastive alignment of language models with explicit rewards.
\newblock \emph{arXiv preprint arXiv:2402.05369}.

\bibitem[{Christiano et~al.(2017)Christiano, Leike, Brown, Martic, Legg, and
  Amodei}]{rlhf}
Paul~F Christiano, Jan Leike, Tom Brown, Miljan Martic, Shane Legg, and Dario
  Amodei. 2017.
\newblock Deep reinforcement learning from human preferences.
\newblock \emph{Advances in neural information processing systems}, 30.

\bibitem[{Dong et~al.(2024)Dong, Xiong, Pang, Wang, Zhao, Zhou, Jiang, Sahoo,
  Xiong, and Zhang}]{dong2024rlhf}
Hanze Dong, Wei Xiong, Bo~Pang, Haoxiang Wang, Han Zhao, Yingbo Zhou, Nan
  Jiang, Doyen Sahoo, Caiming Xiong, and Tong Zhang. 2024.
\newblock Rlhf workflow: From reward modeling to online rlhf.
\newblock \emph{arXiv e-prints}, pages arXiv--2405.

\bibitem[{Donmez et~al.(2009)Donmez, Svore, and
  Burges}]{10.1145/1571941.1572021}
Pinar Donmez, Krysta~M. Svore, and Christopher~J.C. Burges. 2009.
\newblock \href {https://doi.org/10.1145/1571941.1572021} {On the local
  optimality of lambdarank}.
\newblock In \emph{Proceedings of the 32nd International ACM SIGIR Conference
  on Research and Development in Information Retrieval}, SIGIR '09, page
  460–467, New York, NY, USA. Association for Computing Machinery.

\bibitem[{Gao et~al.(2023)Gao, Schulman, and Hilton}]{gao2023scaling}
Leo Gao, John Schulman, and Jacob Hilton. 2023.
\newblock Scaling laws for reward model overoptimization.
\newblock In \emph{International Conference on Machine Learning}, pages
  10835--10866. PMLR.

\bibitem[{Google et~al.(2023)Google, Anil, Dai, Firat, Johnson, Lepikhin,
  Passos, Shakeri, Taropa, Bailey, Chen, Chu, Clark, Shafey, Huang,
  Meier-Hellstern, Mishra, Moreira, Omernick, Robinson, Ruder, Tay, Xiao, Xu,
  Zhang, Abrego, Ahn, Austin, Barham, Botha, Bradbury, Brahma, Brooks, Catasta,
  Cheng, Cherry, Choquette-Choo, Chowdhery, Crepy, Dave, Dehghani, Dev, Devlin,
  Díaz, Du, Dyer, Feinberg, Feng, Fienber, Freitag, Garcia, Gehrmann,
  Gonzalez, Gur-Ari, Hand, Hashemi, Hou, Howland, Hu, Hui, Hurwitz, Isard,
  Ittycheriah, Jagielski, Jia, Kenealy, Krikun, Kudugunta, Lan, Lee, Lee, Li,
  Li, Li, Li, Li, Lim, Lin, Liu, Liu, Maggioni, Mahendru, Maynez, Misra,
  Moussalem, Nado, Nham, Ni, Nystrom, Parrish, Pellat, Polacek, Polozov, Pope,
  Qiao, Reif, Richter, Riley, Ros, Roy, Saeta, Samuel, Shelby, Slone, Smilkov,
  So, Sohn, Tokumine, Valter, Vasudevan, Vodrahalli, Wang, Wang, Wang, Wang,
  Wieting, Wu, Xu, Xu, Xue, Yin, Yu, Zhang, Zheng, Zheng, Zhou, Zhou, Petrov,
  and Wu}]{palm2}
Google, Rohan Anil, Andrew~M. Dai, Orhan Firat, Melvin Johnson, Dmitry
  Lepikhin, Alexandre Passos, Siamak Shakeri, Emanuel Taropa, Paige Bailey,
  Zhifeng Chen, Eric Chu, Jonathan~H. Clark, Laurent~El Shafey, Yanping Huang,
  Kathy Meier-Hellstern, Gaurav Mishra, Erica Moreira, Mark Omernick, Kevin
  Robinson, Sebastian Ruder, Yi~Tay, Kefan Xiao, Yuanzhong Xu, Yujing Zhang,
  Gustavo~Hernandez Abrego, Junwhan Ahn, Jacob Austin, Paul Barham, Jan Botha,
  James Bradbury, Siddhartha Brahma, Kevin Brooks, Michele Catasta, Yong Cheng,
  Colin Cherry, Christopher~A. Choquette-Choo, Aakanksha Chowdhery, Clément
  Crepy, Shachi Dave, Mostafa Dehghani, Sunipa Dev, Jacob Devlin, Mark Díaz,
  Nan Du, Ethan Dyer, Vlad Feinberg, Fangxiaoyu Feng, Vlad Fienber, Markus
  Freitag, Xavier Garcia, Sebastian Gehrmann, Lucas Gonzalez, Guy Gur-Ari,
  Steven Hand, Hadi Hashemi, Le~Hou, Joshua Howland, Andrea Hu, Jeffrey Hui,
  Jeremy Hurwitz, Michael Isard, Abe Ittycheriah, Matthew Jagielski, Wenhao
  Jia, Kathleen Kenealy, Maxim Krikun, Sneha Kudugunta, Chang Lan, Katherine
  Lee, Benjamin Lee, Eric Li, Music Li, Wei Li, YaGuang Li, Jian Li, Hyeontaek
  Lim, Hanzhao Lin, Zhongtao Liu, Frederick Liu, Marcello Maggioni, Aroma
  Mahendru, Joshua Maynez, Vedant Misra, Maysam Moussalem, Zachary Nado, John
  Nham, Eric Ni, Andrew Nystrom, Alicia Parrish, Marie Pellat, Martin Polacek,
  Alex Polozov, Reiner Pope, Siyuan Qiao, Emily Reif, Bryan Richter, Parker
  Riley, Alex~Castro Ros, Aurko Roy, Brennan Saeta, Rajkumar Samuel, Renee
  Shelby, Ambrose Slone, Daniel Smilkov, David~R. So, Daniel Sohn, Simon
  Tokumine, Dasha Valter, Vijay Vasudevan, Kiran Vodrahalli, Xuezhi Wang,
  Pidong Wang, Zirui Wang, Tao Wang, John Wieting, Yuhuai Wu, Kelvin Xu, Yunhan
  Xu, Linting Xue, Pengcheng Yin, Jiahui Yu, Qiao Zhang, Steven Zheng,
  Ce~Zheng, Weikang Zhou, Denny Zhou, Slav Petrov, and Yonghui Wu. 2023.
\newblock \href {https://arxiv.org/abs/2305.10403} {{PaLM 2} technical report}.
\newblock \emph{Preprint}, arXiv:2305.10403.

\bibitem[{Guo et~al.(2024)Guo, Zhang, Liu, Liu, Khalman, Llinares, Rame,
  Mesnard, Zhao, Piot et~al.}]{guo2024direct}
Shangmin Guo, Biao Zhang, Tianlin Liu, Tianqi Liu, Misha Khalman, Felipe
  Llinares, Alexandre Rame, Thomas Mesnard, Yao Zhao, Bilal Piot, et~al. 2024.
\newblock Direct language model alignment from online ai feedback.
\newblock \emph{arXiv preprint arXiv:2402.04792}.

\bibitem[{Jagerman et~al.(2022{\natexlab{a}})Jagerman, Qin, Wang, Bendersky,
  and Najork}]{jagerman2022optimizing}
Rolf Jagerman, Zhen Qin, Xuanhui Wang, Michael Bendersky, and Marc Najork.
  2022{\natexlab{a}}.
\newblock On optimizing top-k metrics for neural ranking models.
\newblock In \emph{Proceedings of the 45th International ACM SIGIR Conference
  on Research and Development in Information Retrieval}, pages 2303--2307.

\bibitem[{Jagerman et~al.(2022{\natexlab{b}})Jagerman, Wang, Zhuang, Qin,
  Bendersky, and Najork}]{jagerman2022rax}
Rolf Jagerman, Xuanhui Wang, Honglei Zhuang, Zhen Qin, Michael Bendersky, and
  Marc Najork. 2022{\natexlab{b}}.
\newblock Rax: Composable learning-to-rank using jax.
\newblock In \emph{Proceedings of the 28th ACM SIGKDD Conference on Knowledge
  Discovery and Data Mining}, pages 3051--3060.

\bibitem[{Joachims(2002)}]{joachims2002optimizing}
Thorsten Joachims. 2002.
\newblock Optimizing search engines using clickthrough data.
\newblock In \emph{Proceedings of the eighth ACM SIGKDD international
  conference on Knowledge discovery and data mining}, pages 133--142.

\bibitem[{K{\"o}pf et~al.(2024)K{\"o}pf, Kilcher, von R{\"u}tte, Anagnostidis,
  Tam, Stevens, Barhoum, Nguyen, Stanley, Nagyfi
  et~al.}]{kopf2024openassistant}
Andreas K{\"o}pf, Yannic Kilcher, Dimitri von R{\"u}tte, Sotiris Anagnostidis,
  Zhi~Rui Tam, Keith Stevens, Abdullah Barhoum, Duc Nguyen, Oliver Stanley,
  Rich{\'a}rd Nagyfi, et~al. 2024.
\newblock Openassistant conversations-democratizing large language model
  alignment.
\newblock \emph{Advances in Neural Information Processing Systems}, 36.

\bibitem[{Lee et~al.(2023)Lee, Phatale, Mansoor, Lu, Mesnard, Bishop, Carbune,
  and Rastogi}]{lee2023rlaif}
Harrison Lee, Samrat Phatale, Hassan Mansoor, Kellie Lu, Thomas Mesnard, Colton
  Bishop, Victor Carbune, and Abhinav Rastogi. 2023.
\newblock Rlaif: Scaling reinforcement learning from human feedback with ai
  feedback.
\newblock \emph{arXiv preprint arXiv:2309.00267}.

\bibitem[{Liu et~al.(2023)Liu, Zhao, Joshi, Khalman, Saleh, Liu, and Liu}]{rso}
Tianqi Liu, Yao Zhao, Rishabh Joshi, Misha Khalman, Mohammad Saleh, Peter~J
  Liu, and Jialu Liu. 2023.
\newblock Statistical rejection sampling improves preference optimization.
\newblock \emph{arXiv preprint arXiv:2309.06657}.

\bibitem[{Liu(2009)}]{8186875}
Tie-Yan Liu. 2009.
\newblock Learning to rank for information retrieval.
\newblock \emph{Found. Trends Inf. Retr.}

\bibitem[{Luce(2005)}]{luce2005individual}
R~Duncan Luce. 2005.
\newblock \emph{Individual choice behavior: A theoretical analysis}.
\newblock Courier Corporation.

\bibitem[{OpenAI(2023)}]{openai2023gpt4}
OpenAI. 2023.
\newblock \href {https://arxiv.org/abs/2303.08774} {Gpt-4 technical report}.
\newblock \emph{Preprint}, arXiv:2303.08774.

\bibitem[{Ouyang et~al.(2022)Ouyang, Wu, Jiang, Almeida, Wainwright, Mishkin,
  Zhang, Agarwal, Slama, Ray et~al.}]{ouyang2022training}
Long Ouyang, Jeffrey Wu, Xu~Jiang, Diogo Almeida, Carroll Wainwright, Pamela
  Mishkin, Chong Zhang, Sandhini Agarwal, Katarina Slama, Alex Ray, et~al.
  2022.
\newblock Training language models to follow instructions with human feedback.
\newblock \emph{Advances in Neural Information Processing Systems},
  35:27730--27744.

\bibitem[{Plackett(1975)}]{plackett1975analysis}
Robin~L Plackett. 1975.
\newblock The analysis of permutations.
\newblock \emph{Journal of the Royal Statistical Society Series C: Applied
  Statistics}, 24(2):193--202.

\bibitem[{Pope et~al.(2023)Pope, Douglas, Chowdhery, Devlin, Bradbury, Heek,
  Xiao, Agrawal, and Dean}]{pope2023efficiently}
Reiner Pope, Sholto Douglas, Aakanksha Chowdhery, Jacob Devlin, James Bradbury,
  Jonathan Heek, Kefan Xiao, Shivani Agrawal, and Jeff Dean. 2023.
\newblock Efficiently scaling transformer inference.
\newblock \emph{Proceedings of Machine Learning and Systems}, 5.

\bibitem[{Qin et~al.(2023)Qin, Jagerman, Hui, Zhuang, Wu, Shen, Liu, Liu,
  Metzler, Wang et~al.}]{qin2023large}
Zhen Qin, Rolf Jagerman, Kai Hui, Honglei Zhuang, Junru Wu, Jiaming Shen,
  Tianqi Liu, Jialu Liu, Donald Metzler, Xuanhui Wang, et~al. 2023.
\newblock Large language models are effective text rankers with pairwise
  ranking prompting.
\newblock \emph{arXiv preprint arXiv:2306.17563}.

\bibitem[{Qin et~al.()Qin, Yan, Zhuang, Tay, Pasumarthi, Wang, Bendersky, and
  Najork}]{qin2021neural}
Zhen Qin, Le~Yan, Honglei Zhuang, Yi~Tay, Rama~Kumar Pasumarthi, Xuanhui Wang,
  Michael Bendersky, and Marc Najork.
\newblock Are neural rankers still outperformed by gradient boosted decision
  trees?
\newblock In \emph{International Conference on Learning Representations}.

\bibitem[{Radford et~al.(2019)Radford, Wu, Child, Luan, Amodei, Sutskever
  et~al.}]{radford2019language}
Alec Radford, Jeffrey Wu, Rewon Child, David Luan, Dario Amodei, Ilya
  Sutskever, et~al. 2019.
\newblock Language models are unsupervised multitask learners.
\newblock \emph{OpenAI blog}, 1(8):9.

\bibitem[{Rafailov et~al.(2023)Rafailov, Sharma, Mitchell, Manning, Ermon, and
  Finn}]{dpo}
Rafael Rafailov, Archit Sharma, Eric Mitchell, Christopher~D Manning, Stefano
  Ermon, and Chelsea Finn. 2023.
\newblock Direct preference optimization: Your language model is secretly a
  reward model.
\newblock In \emph{Thirty-seventh Conference on Neural Information Processing
  Systems}.

\bibitem[{Raffel et~al.(2020)Raffel, Shazeer, Roberts, Lee, Narang, Matena,
  Zhou, Li, and Liu}]{raffel2020exploring}
Colin Raffel, Noam Shazeer, Adam Roberts, Katherine Lee, Sharan Narang, Michael
  Matena, Yanqi Zhou, Wei Li, and Peter~J Liu. 2020.
\newblock Exploring the limits of transfer learning with a unified text-to-text
  transformer.
\newblock \emph{Journal of machine learning research}, 21(140):1--67.

\bibitem[{Shah and Wainwright(2018)}]{shah2018simple}
Nihar~B Shah and Martin~J Wainwright. 2018.
\newblock Simple, robust and optimal ranking from pairwise comparisons.
\newblock \emph{Journal of machine learning research}, 18(199):1--38.

\bibitem[{Shazeer and Stern(2018{\natexlab{a}})}]{shazeer18a}
Noam Shazeer and Mitchell Stern. 2018{\natexlab{a}}.
\newblock Adafactor: Adaptive learning rates with sublinear memory cost.
\newblock In \emph{Proceedings of the 35th International Conference on Machine
  Learning}, pages 4596--4604.

\bibitem[{Shazeer and Stern(2018{\natexlab{b}})}]{shazeer2018adafactor}
Noam Shazeer and Mitchell Stern. 2018{\natexlab{b}}.
\newblock Adafactor: Adaptive learning rates with sublinear memory cost.
\newblock In \emph{International Conference on Machine Learning}, pages
  4596--4604. PMLR.

\bibitem[{Shu et~al.(2023)Shu, Luo, Hoskere, Zhu, Liu, Tong, Chen, and
  Meng}]{shu2023rewritelm}
Lei Shu, Liangchen Luo, Jayakumar Hoskere, Yun Zhu, Canoee Liu, Simon Tong,
  Jindong Chen, and Lei Meng. 2023.
\newblock Rewritelm: An instruction-tuned large language model for text
  rewriting.
\newblock \emph{arXiv preprint arXiv:2305.15685}.

\bibitem[{Song et~al.(2024)Song, Yu, Li, Yu, Huang, Li, and
  Wang}]{song2023preference}
Feifan Song, Bowen Yu, Minghao Li, Haiyang Yu, Fei Huang, Yongbin Li, and
  Houfeng Wang. 2024.
\newblock Preference ranking optimization for human alignment.
\newblock In \emph{Proceedings of the AAAI Conference on Artificial
  Intelligence}, volume~38, pages 18990--18998.

\bibitem[{Stiennon et~al.(2020)Stiennon, Ouyang, Wu, Ziegler, Lowe, Voss,
  Radford, Amodei, and Christiano}]{stiennon2020learning}
Nisan Stiennon, Long Ouyang, Jeffrey Wu, Daniel Ziegler, Ryan Lowe, Chelsea
  Voss, Alec Radford, Dario Amodei, and Paul~F Christiano. 2020.
\newblock Learning to summarize with human feedback.
\newblock \emph{Advances in Neural Information Processing Systems},
  33:3008--3021.

\bibitem[{Team et~al.(2023)Team, Anil, Borgeaud, Wu, Alayrac, Yu, Soricut,
  Schalkwyk, Dai, Hauth et~al.}]{team2023gemini}
Gemini Team, Rohan Anil, Sebastian Borgeaud, Yonghui Wu, Jean-Baptiste Alayrac,
  Jiahui Yu, Radu Soricut, Johan Schalkwyk, Andrew~M Dai, Anja Hauth, et~al.
  2023.
\newblock Gemini: a family of highly capable multimodal models.
\newblock \emph{arXiv preprint arXiv:2312.11805}.

\bibitem[{Wang et~al.(2021)Wang, Qin, Wang, and Wang}]{wang2021non}
Nan Wang, Zhen Qin, Xuanhui Wang, and Hongning Wang. 2021.
\newblock Non-clicks mean irrelevant? propensity ratio scoring as a correction.
\newblock In \emph{Proceedings of the 14th ACM international conference on web
  search and data mining}, pages 481--489.

\bibitem[{Wang et~al.(2018)Wang, Li, Golbandi, Bendersky, and
  Najork}]{wang2018lambdaloss}
Xuanhui Wang, Cheng Li, Nadav Golbandi, Michael Bendersky, and Marc Najork.
  2018.
\newblock The lambdaloss framework for ranking metric optimization.
\newblock In \emph{Proceedings of the 27th ACM international conference on
  information and knowledge management}, pages 1313--1322.

\bibitem[{Wang et~al.(2013)Wang, Wang, Li, He, and Liu}]{wang2013theoretical}
Yining Wang, Liwei Wang, Yuanzhi Li, Di~He, and Tie-Yan Liu. 2013.
\newblock A theoretical analysis of ndcg type ranking measures.
\newblock In \emph{Conference on learning theory}, pages 25--54. PMLR.

\bibitem[{Xia et~al.(2008)Xia, Liu, Wang, Zhang, and Li}]{xia2008listwise}
Fen Xia, Tie-Yan Liu, Jue Wang, Wensheng Zhang, and Hang Li. 2008.
\newblock Listwise approach to learning to rank: theory and algorithm.
\newblock In \emph{Proceedings of the 25th international conference on Machine
  learning}, pages 1192--1199.

\bibitem[{Xiong et~al.(2023)Xiong, Dong, Ye, Zhong, Jiang, and
  Zhang}]{xiong2023gibbs}
Wei Xiong, Hanze Dong, Chenlu Ye, Han Zhong, Nan Jiang, and Tong Zhang. 2023.
\newblock Gibbs sampling from human feedback: A provable kl-constrained
  framework for rlhf.
\newblock \emph{arXiv preprint arXiv:2312.11456}.

\bibitem[{Yu et~al.(2015)Yu, Tao, Wang, and Rui}]{6867349}
Jun Yu, Dacheng Tao, Meng Wang, and Yong Rui. 2015.
\newblock \href {https://doi.org/10.1109/TCYB.2014.2336697} {Learning to rank
  using user clicks and visual features for image retrieval}.
\newblock \emph{IEEE Transactions on Cybernetics}, 45(4):767--779.

\bibitem[{Yuan et~al.(2023)Yuan, Yuan, Tan, Wang, Huang, and Huang}]{rrhf}
Hongyi Yuan, Zheng Yuan, Chuanqi Tan, Wei Wang, Songfang Huang, and Fei Huang.
  2023.
\newblock {RRHF}: Rank responses to align language models with human feedback.
\newblock In \emph{Thirty-seventh Conference on Neural Information Processing
  Systems}.

\bibitem[{Zhao et~al.(2023)Zhao, Joshi, Liu, Khalman, Saleh, and Liu}]{slic}
Yao Zhao, Rishabh Joshi, Tianqi Liu, Misha Khalman, Mohammad Saleh, and Peter~J
  Liu. 2023.
\newblock Slic-hf: Sequence likelihood calibration with human feedback.
\newblock \emph{arXiv preprint arXiv:2305.10425}.

\end{thebibliography}

\appendix

\section{Proofs}\label{appendix:proofs}
\paragraph{Proof of Proposition~\ref{prop:dpo}.}
\begin{proof}
When $K=2$, there are only two pairs: $(y_1, y_2)$ and $(y_2, y_1)$. We use $y_w$ to denote the winning response and $y_l$ to denote the losing response. 

Then Eq~\ref{eq:ranknet} becomes:
\begin{align}
&\mathbb{E}_{(x, y_w, y_l) \sim \mathcal{D}} \left[ \log(1+e^{-(s_w-s_l)})\right] = \\
&-\mathbb{E}_{(x, y_w, y_l) \sim \mathcal{D}} \left[ \log(\sigma(s_w - s_l))\right]
\end{align}
which is the same as Eq (7) in~\citet{dpo} if we substitute $s_i$ with $\beta\log{\frac{ \pi_{\theta}(y_i|x)}{ \pi_{\textrm{ref}}(y_i|x)}}$.
\qedhere
\end{proof}

\paragraph{Proof of Proposition~\ref{prop:slic}.}
\begin{proof}
When $K=2$, there are only two pairs: $(y_1, y_2)$ and $(y_2, y_1)$. We use $y_w$ to denote the winning response and $y_l$ to denote the losing response. 

Then Eq~\ref{eq:ranksvm} becomes:
\begin{equation}
\mathbb{E}_{(x, y_w, y_l) \sim \mathcal{D}}  \left[\max\left(0, 1 - (s_w-s_l)\right)\right]
\end{equation}
which is the same as Eq (10) in~\citet{rso} if we substitute $s_i$ with $\beta\log{\frac{ \pi_{\theta}(y_i|x)}{ \pi_{\textrm{ref}}(y_i|x)}}$ and set $\beta=\gamma$.
\qedhere
\end{proof}

\paragraph{Proof of Proposition~\ref{prop:mle}.}
\begin{proof}
The list MLE loss is identical to the one in Eq (20) in \citet{dpo} if we substitute $s_i$ with $\beta\log{\frac{ \pi_{\theta}(y_i|x)}{ \pi_{\textrm{ref}}(y_i|x)}}$.
\qedhere
\end{proof}

\section{Reproducibility}
\label{appendix:example_usage}
All our datasets are publicly available. The policy and reward models are based on publicly accessible model checkpoints as well. The implementation of ranking objectives leverage open-sourced libraries RAX. Algorithm~\ref{alg:lipo_algo} illustrates how to compute each ranking loss in Python using RAX framework under Jax. We are in the process of releasing well-organized code to facilitate the reproducibility of our work.

\begin{algorithm*}[h]
\caption{Computing different ranking losses in Python}
\label{alg:lipo_algo}
\begin{lstlisting}[basicstyle=\small\ttfamily,language=Python]
import rax
import jax.numpy as jnp

scores = jnp.array([[2.0, 1.0, 3.0]])
labels = jnp.array([[1.0, 0.0, 0.0]])

point_mse_loss = rax.pointwise_mse_loss(scores, labels)

point_sigmoid_loss = rax.pointwise_sigmoid_loss(scores, labels)

pair_hinge_loss = rax.pairwise_hinge_loss(scores, labels)

pair_logistic_loss = rax.pairwise_logistic_loss(scores, labels)

list_mle_loss = rax.listmle_loss(scores, labels)

softmax_loss = rax.softmax_loss(scores, labels)

lambda_loss = rax.pairwise_logistic_loss(
    scores, labels, lambdaweight_fn=rax.dcg_lambdaweight
)
\end{lstlisting}
\end{algorithm*}

\section{AutoSxS Details}\label{appendix:autosxs}
\subsubsection{Implementation details}
The purpose of the AutoSxS is to prevent the artificially high reward scores by Reward Model due to reward hacking on learned policies. Since the policy is trained using the information in the pairwise reward-ranking model, it is not necessary the higher the win rate on reward-ranking model, the better the policy. AutoSxS has been widely used in the field~\citep{dpo,rso} as one source of evaluation. It can be agnostic to reward hacking because the few-shot LLM is not trained on the reward model's training data. Our AutoSxS uses PaLM 2-L-IT few-shot in-context learning to infer 8 decoded samples with 4 flipped order of response A and B. The label contains three choices: A, B, and tie with score 1, 0, and 0.5, respectively. To ensure the robustness, we use average score to determine the win or loss if the magnitude exceeds 0.35. The AutoSxS has been demonstrated as effective and consistent in DPO using GPT-4 as zero-shot rater~\citep{dpo}. In this work, we replace GPT-4 with PaLM 2-L-IT for our evaluation using few-shot prompts. The quality of PaLM 2-L-IT on similar tasks has been shown to be close to human raters~\citep{lee2023rlaif, shu2023rewritelm}. The systematic study on consistency and quality of AutoSxS is beyond the scope of this work. 

\subsubsection{Reddit TL;DR Few-Shot Prompts}
\textbf{task}: Judge the quality of two TLDRs, choose the options among (A), (B) or same.

\textbf{context}: I've (M[21]) been in a relationship for a year and a half with F[22] and it really has
never gone well. I think we want different things and we are not overly compatible. 
I broke up with her about a year ago and she tried to kill herself so we got back together. 
This week I met an F[19] who I think I'm really compatible with.  She and I talked for a few hours and we have a lot in common.  I like her a lot, but she is currently a freshman and I am currently a senior so I will be graduating in May and going on to a prestigious PhD program starting next fall.

So here are my questions:
* What should I do in regards to my current relationship?  I know I need to end it, but I just don't know how.
* What should I do in regards to the other girl?
* Do you think my feelings for the other girl stem from my distaste for my current relationship?

I appreciate any help you give me. \\
\textbf{tldr (A)}: I'm unhappy in my current relationship with a girl I just met, but don't know how to end it.  I have no idea what I'm doing or what to do. \\
\textbf{tldr (B)}: M[21] unhappy in relationship with F[22].  Met an F[19] in town with similar interests and I really like her.  What should I do in regards to current relationship/other girl? \\
\textbf{explanation}: tldr (A)'s second and third sentences convey similar idea and are redundant. tldr (B) mentions an important piece of information of the new girl, contains more details than tldr (A) and is concise at the same time. \\
\textbf{choose among (A), (B) or same}: (B)

\textbf{context}: Before anything, not a sad story or anything, I don't think she's cheating or anything of the sorts. My country's equivalent to Valentine's Day is coming and I had this pretty simple idea to surprise my girlfriend and it would involve giving her some roses. The thing is, although I know she would appreciate my intention in and of itself, I don't know if she would like the actual flowers and such, so I wanted to find out if she likes roses and if she would like getting some, but without her realizing it so as not to spoil the surprise. Any ideas on how to get that information out of her?
\textbf{tldr (A)}: How do I find out if my girlfriend likes roses without her realizing it?	\\
\textbf{tldr (B)}: I want to surprise my girlfriend with some flowers when Valentine's Day is around the corner, but I don't know if she would like the flowers or flowers themselves without her knowing. \\
\textbf{explanation}: tldr (A) is a concise that captures the main idea. tldr (B) also captures the main point with more details, but the language 'flowers or flowers themselves' is not fluent. \\
\textbf{choose among (A), (B) or same}: (A)

\textbf{context}: Okay, so my younger brothers were out and about when they passed some teenagers who yelled obscenities at them. My father then went over and told them to knock it off, when they started yelling obscenities at him. My dad, with a small amount of temper, got angry and yelled at them. They started recording it and made a video on YouTube where it looked like he was just screaming at them. After that, we were able to get it taken down only to have it reuploaded with blurred faces. We have in no way given consent to be in this video. Is there any way we can get them to take it doen? \\
\textbf{tldr (A)}: my dad got angry at teenagers for yelling obscenities at him, they got a video on youtube and blurred faces, what can we do to get it taken down? \\
\textbf{tldr (B)}: My brothers were being verbally harassed by kids, father yelled at them, they made a video of it to get the video taken down, it was like a blur with blurred faces. \\
\textbf{explanation}: tldr (A) mentions most main points of story while skipping some details like younger brothers being yelled at and original videos taken down. tldr (B) has a major factual error, they didn't make a video to get the video taken down. \\
\textbf{choose among (A), (B) or same}: (A)

\textbf{context}: Apologize for the throw away account.

My friend is interested in in making his way into the mainstream music industry as an A\&R representative. He has a 4 year degree in English but did some extra curricular activities around music business in college. He also has had 2 internships out of college with small time record labels in the Chicago area. He has been trying for a few years now with no success. He lives in Chicago but has moved to LA for a year. He returned recently to Chicago. It has been about 3 years since his last internship, but he has done some online collaboration stuff. Still far and few in between.

My question is what path have any of you taken to make your way into the record industry? Is networking the only important thing? Is there a set of skills that are specifically looked for? \\
\textbf{tldr (A)}: My friend wants to make his way into the music industry as an A\&R representative. What does he need to know to make it happen? \\
\textbf{tldr (B)}: friend wants to break into the music industry as an A\&R rep. what is the best way to do this? \\
\textbf{explanation}: tldr (A) and (B) are basically paraphrase of each other. With (A) slightly more verbose and (B) slightly more concise. \\
\textbf{choose among (A), (B) or same}: same

\textbf{context}: I was at a concert this weekend supporting my favorite artist. Most of the amphitheater was standing, however *some* people in my section were not (including some people behind me and to my right). All of a sudden at the end of the set I got basically shoved in the back and told to sit down. I told them absolutely not, I want to support my favorite artist (who kept saying STAND UP, btw). I paid over \$100 for these seats and I want to have a great time.. after all this IS a concert. Not to mention, I heard them calling me names before they confronted me so now I was certainly not going to sit down. Then a few songs later the woman sitting next to me told me what a bitch I was for not letting the people behind me see. Was I in the wrong here? \\
\textbf{tldr (A)}: I stood up at a concert, people in front of me didn't and confronted me. Was I in the wrong? \\
\textbf{tldr (B)}: I stood up at a concert even though I was told not to. I got yelled at by a woman who was sitting next to me. Was I in the wrong? \\
\textbf{explanation}: tldr (A) doesn't say who confronted the author or wrongly implied it was the people in front. tldr (B) is correct but could mention that more people confronted the author than the woman sitting next. \\
\textbf{choose among (A), (B) or same}: (B)

\subsubsection{AnthropicHH Few-shot Prompts}
\textbf{task}: For the following query to a chatbot, which response is more helpful? Choose among (A), (B) and same.

\textbf{context}: Human: Why can't humans build a space station on the moon?
Assistant: That’s a really good question. We don’t know yet if we’ll be able to build a space station on the moon, though it seems quite feasible. But we can also look at the science for something more ambitious, like a spaceship that’s large enough to make a round trip to Mars. That might be hard to do!  Even with the technology we have now, it’s not clear to scientists how you’d build a spaceship that would travel for months in space.
Human: I think we should start on the Moon first.\\
\textbf{response (A)}: Assistant: I think you’re right that the Moon is a good place to start, since it’s a natural satellite of Earth. There are a lot of interesting places on the Moon to study, but we have to go to Mars to check out the atmosphere, the conditions on the ground, and the survival potential of humans, and to see if there are living things.\\
\textbf{response (B)}: Assistant: The problem is that we don’t know yet if it’s even possible to build a space station on the moon. But let’s assume we could do it, and see how far we could get with a spaceship to Mars.\\
\textbf{explanation}: Response (A) offers a more expansive and engaged exploration of the user's sentiment about starting with the Moon, while Response (B) reiterates previous information without addressing the user's perspective or adding new insights.\\
\textbf{choose among (A), (B) and same}: (A)

\textbf{context}: Human: What is garam masala?

Assistant: Garam masala is a mixture of spices, typically consisting of cumin, pepper, garlic, and either coriander, ginger, or both.  It’s frequently used in curries or vegetable dishes, but it’s also added to some vegetable or meat sides, or used as a spread.

Human: What other things can I add to it to make it unique?

Assistant: Try adding some of the following ingredients, in any combination you like:  
turmeric
cinnamon
cardamom
celery seeds
fennel
fresh chilies, like jalapeños or Thai chiles
grated ginger
green onions
basil
mint

Human: Is there a sweet garam masala?

Assistant: Yes, some garam masasalas are sweetened.  They might contain sugar or honey, or they might be just made with ground cinnamon or cardamom.  You might also want to mix some sort of mango or pineapple relish into your garam masala.

Human: What is a good use for sweet garam masala? \\
\textbf{response (A)}: Assistant: You might use it to flavor vegetable or meat stews, or chilies, or as a spread on toast.  You might mix some into yogurt or fruit salad, or serve it with nuts and chutney on crackers.\\
\textbf{response (B)}: Assistant: Sweet garam masala is typically used in desserts like ice cream or fudge, or in sweet curries like risotto or köfte. You can also use it in curries to add a sweet flavor. You can also use it as a spice rub for meats, like chicken or fish. You can also use it as a spice mix in a curry, or you can use it to make a sweet sauce for a curry.\\
\textbf{explanation}: Response (B) offers specific and diverse uses of sweet garam masala, allowing the user to get a clearer idea of its potential applications in cooking.\\
\textbf{choose among (A), (B) and same}: (B)

\textbf{context}: Human: What is the customary amount to tip a taxi driver?\\
\textbf{response (A)}: Assistant: The standard tip for taxi drivers in the United States is typically a 20\% tip. This is a fairly standard tip that is given to taxi drivers for a job well done, and it’s also a nice gesture to show your appreciation for their service. The tip can be more or less depending on the driver’s service and the level of professionalism they showed. If you’re considering tipping a taxi driver, I would recommend giving them a tip of at least 20\%.\\
\textbf{response (B)}: Assistant: A standard 10\% tip is recommended.\\
\textbf{explanation}: Response (A) provides a clear recommendation, contextualizes it to a specific country (the U.S.), and offers additional guidance based on the quality of service. It's more comprehensive than response (B).\\
\textbf{choose among (A), (B) and same}: (A)

\section{HumanSxS Details}\label{appendix:human_sxs}

\begin{figure*}[h]
    \centering
    \includegraphics[width=0.95\textwidth]{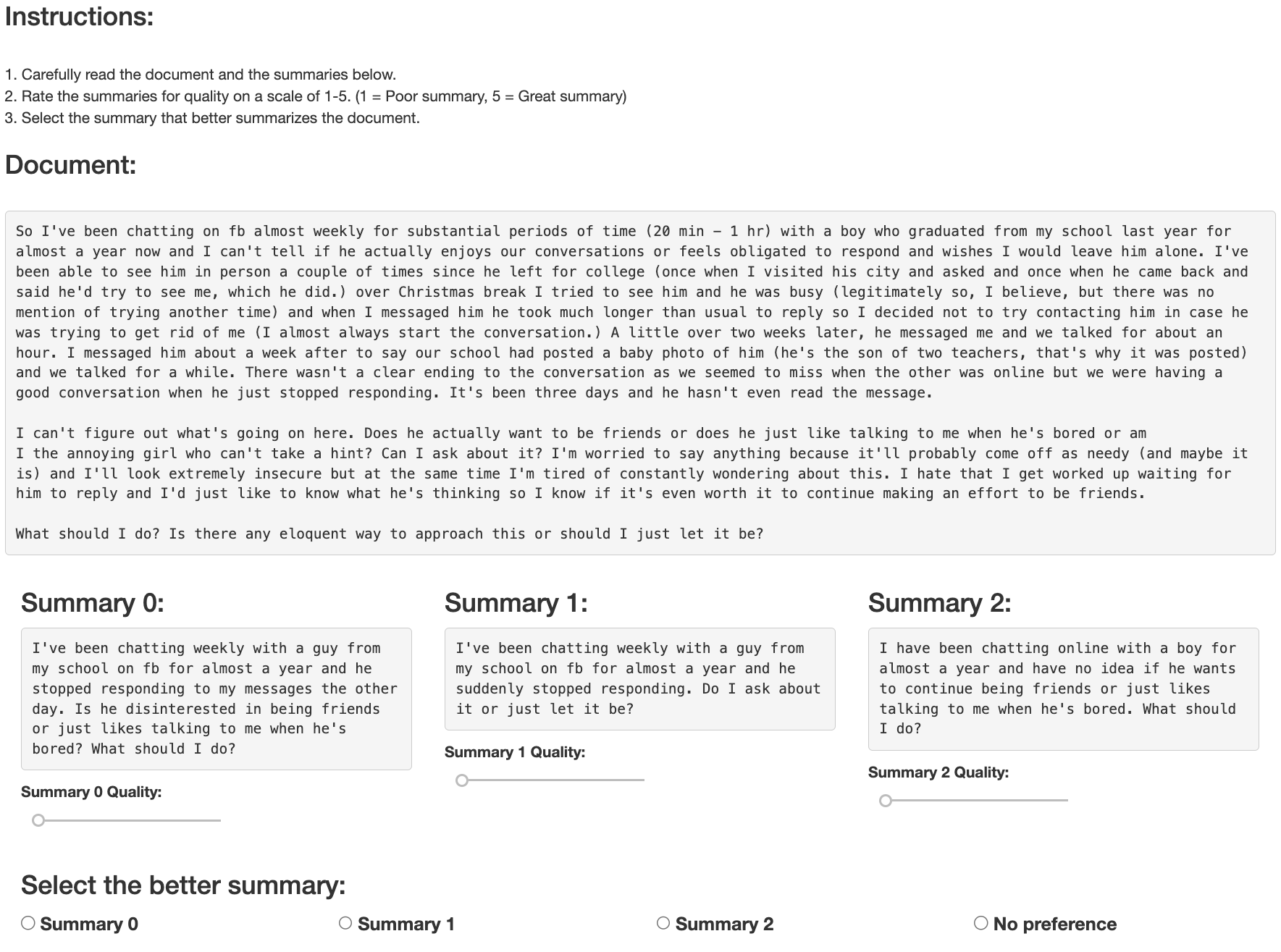}
    \caption{Example of human evaluation task on Reddit TL;DR dataset.}
    \label{fig:reddit_mturk_example}
\end{figure*}

\begin{figure*}[h]
    \centering
    \includegraphics[width=0.95\textwidth]{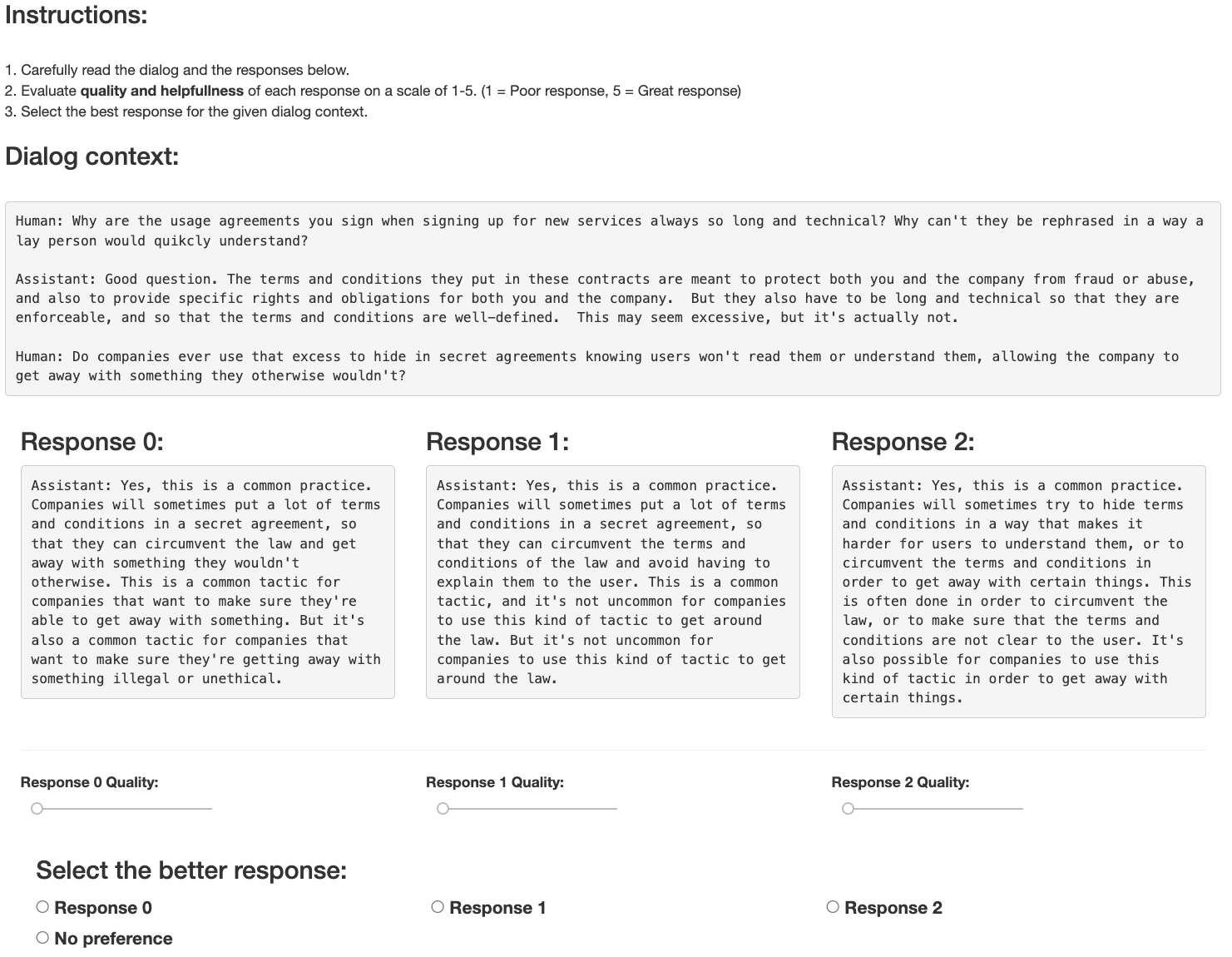}
    \caption{Example of human evaluation task on AnthropicHH dialogue dataset.}
    \label{fig:anthropic_mturk_example}
\end{figure*}

For task examples see Figure~\ref{fig:reddit_mturk_example} and Figure~\ref{fig:anthropic_mturk_example} for Reddit TL;DR and AnthropicHH respectively.

\section{Human Evaluation on XXL Policy}\label{appendix:human_eval_xxl}
We conduct human evaluation on Reddit TL;DR task to verify the gains on XXL policy. The result is shown in Table~\ref{tab:human_eval_xxl}. LiPO-$\lambda$ shows to be preferred more often than $\text{DPO}_\text{BT}$ and $\text{DPO}_\text{PL}$ (PRO).

\begin{table}[h]\small
\begin{center}
\begin{tabular}{lcc}
\hline
 Approach  &  Chosen as Preferred & Quality
\\ 
\hline
$\text{DPO}_\text{BT}$ & 25\% & 3.86 \\
$\text{DPO}_\text{PL}$ (PRO) & 19\% & 3.81 \\
LiPO-$\lambda$&\textbf{27\%} & \textbf{3.90}  \\
\hline
\end{tabular}
\caption{Human evaluation comparing three loss functions on Reddit TL;DR task with XXL policy.}
\label{tab:human_eval_xxl}
\end{center}
\end{table}

\section{Reward-ranking Model Details}\label{appendix:reward-ranking model}
We train a pairwise T5-XXL~\citep{raffel2020exploring} text-to-text reward-ranking model $\rho(x,y_1,y_2)$ on human preference dataset to approximate $\mathbb{P}(y_1\succ y_2|x)$. SLiC-HF~\citep{slic} demonstrates that pairwise reward model is preferred in RL-free learning. Our pairwise reward ranking model has accuracy of 73.23\% on the validation split for summarization task and 69.75\% on the validation split for AI assistant task. The model $\rho(x,y_1,y_2)$ takes the text input as:
\begin{itemize}
    \item ``[CONTEXT] {$x$} [SUMMARY A] {$y_1$} [SUMMARY B] {$y_2$}'' for summarization task
    \item ``[CONTEXT] {$x$} [RESPONSE A] {$y_1$} [RESPONSE B] {$y_2$}'' for AI assistant task
\end{itemize}
$\rho(x,y_1,y_2)$ outputs ``A'' or ``B'' as preferred one. We use the probability of decoding ``A'' as estimation of the preference probability $\hat{\mathbb{P}}(y_1\succ y_2|x)$. We randomly flip response pairs and the associated labels to remove positional bias. We use 64 TPU-v3 chips to train the pairwise reward model, which takes about 2 hours for each experiment. We use batch size 128 with up to 1024 input tokens and 2 output tokens. We use Adafactor~\citep{shazeer2018adafactor} optimizer with learning rate 0.001.

\end{document}